\documentclass[10pt,journal,compsoc]{IEEEtran}
\ifCLASSOPTIONcompsoc
  \usepackage[nocompress]{cite}
\else
  \usepackage{cite}
\fi
\ifCLASSINFOpdf
\else
\fi
\bstctlcite{IEEEexample:BSTcontrol}
\usepackage{algpseudocode}
\usepackage{todonotes}
\usepackage{amsmath,amssymb,amsfonts}
\newcommand{\Lim}[1]{\raisebox{0.5ex}{\scalebox{0.8}{$\displaystyle \lim_{#1}\;$}}}
\usepackage{graphicx}
\usepackage{subcaption}
\usepackage{url}
\usepackage{booktabs}
\DeclareMathOperator*{\argmin}{arg\,min}
\usepackage{color, colortbl}
\definecolor{Gray}{gray}{0.9}
\usepackage{array}
\newcolumntype{L}[1]{>{\raggedright\let\newline\\\arraybackslash\hspace{0pt}}m{#1}}
\usepackage{setspace} 
\usepackage{enumitem}
\usepackage{makecell} %
\usepackage{algorithm}
\usepackage{breakurl}
\usepackage[breaklinks]{hyperref}

\begin{document}
\bstctlcite{IEEEexample:BSTcontrol}
%
\title{Defending Regression Learners Against Poisoning Attacks}
%
%
%
%

\author{Sandamal Weerasinghe, Sarah M. Erfani, Tansu Alpcan, Christopher Leckie and Justin Kopacz
\IEEEcompsocitemizethanks{\IEEEcompsocthanksitem S. Weerasinghe and T. Alpcan are with the Department of Electrical and Electronic Engineering, University of Melbourne, Parkville 3010 Victoria, Australia.\protect\\
E-mail: pweerasinghe@student.unimelb.edu.au
\IEEEcompsocthanksitem S. M. Erfani and C. Leckie are with the School of Computing and Information Systems, University of Melbourne, Parkville 3010 Victoria, Australia.\protect
\IEEEcompsocthanksitem J. Kopacz is with the Northrop Grumman Corporation, USA.}
}

%
%

\markboth{Journal of \LaTeX\ Class Files,~Vol.~14, No.~8, August~2015}%
{Shell \MakeLowercase{\textit{et al.}}: Bare Demo of IEEEtran.cls for Computer Society Journals}
%



\IEEEtitleabstractindextext{%
\begin{abstract}
Regression models, which are widely used from engineering applications to financial forecasting, are vulnerable to targeted malicious attacks such as training data poisoning, through which adversaries can manipulate their predictions. Previous works that attempt to address this problem rely on assumptions about the nature of the attack/attacker or overestimate the knowledge of the learner, making them impractical. We introduce a novel Local Intrinsic Dimensionality (LID) based measure called N-LID that measures the local deviation of a given data point's LID with respect to its neighbors. We then show that N-LID can distinguish poisoned samples from normal samples and propose an N-LID based defense approach that makes no assumptions of the attacker. Through extensive numerical experiments with benchmark datasets, we show that the proposed defense mechanism outperforms the state of the art defenses in terms of prediction accuracy (up to $76\%$ lower MSE compared to an undefended ridge model) and running time.
\end{abstract}

}

\markboth{}{}
\maketitle

\IEEEdisplaynontitleabstractindextext

%
\IEEEpeerreviewmaketitle

\IEEEraisesectionheading{\section{Introduction}\label{sec:introduction}}
\IEEEPARstart{L}{inear} regression models are a fundamental class of supervised learning, with applications in healthcare, business, security, and engineering \cite{koh2011data,bose2001business,naseem2010linear, vrablecova2018smart}. Recent works in the literature show that the performance of regression models degrades significantly in the presence of poisoned training data \cite{jagielski2018manipulating,liu2017robust,xiao2015feature}. Through such attacks, adversaries attempt to force the learner to end up with a prediction model with impaired prediction capabilities. Any application that relies on regression models for automated decision making could potentially be compromised and make decisions that could have serious consequences. 

For example, as demonstrated by Vrablecov{\'a} et al. \cite{vrablecova2018smart}, regression models can be used for forecasting the power load in a smart electrical grid. In such a setting, if an adversary compromises that forecasting system and forces it to predict a lower power demand than the expected power demand for a particular period, the power supplied during that period would be insufficient and may cause blackouts. Conversely, if the adversary forces the forecasting system to predict a higher power demand than the expected demand, there would be surplus power generated that may overload the power distribution system.

In practice, such attacks can take place in situations where the attacker has an opportunity to introduce poisoned samples to the training process. For example, this can occur when data is collected using crowd-sourcing marketplaces, where organizations build data sets with the help of individuals whose authenticity cannot be guaranteed. Due to the size and complexity of datasets, the sponsoring organization may not be able to extensively validate the quality of all data/labels. Therefore, to address this problem, defense mechanisms that take adversarial perturbations into account need to be embedded into regression models. 

Although adversarial manipulations pose a significant threat to critical applications of regression, only a few previous studies have attempted to address this problem. Most works in the literature are related to robust regression, where regression models are trained in the presence of stochastic noise instead of maliciously poisoned data \cite{huang2015robust, mangasarian2000robust}. Recently, however, several works have presented linear regression models that consider the presence of adversarial data samples. Jagielski et al. \cite{jagielski2018manipulating} and Liu et al. \cite{liu2017robust} present two such defense models that iteratively exclude data samples with the largest residuals from the training process. However, both require the learner to be aware of the number of normal samples in the training dataset, which can be considered as an overestimation of the learner's knowledge, making them impractical.

Local Intrinsic Dimensionality (LID) is a metric known for its capability to characterize adversarial examples \cite{LID1_Houle,LID2_Houle}. LID has been applied for detecting adversarial samples in Deep Neural Networks (DNNs) \cite{LID_sarah_ICLR} and as a mechanism to reduce the effect of noisy labels for training DNNs \cite{pmlr-v80-ma18d}. In this paper, we propose a novel LID-based defense mechanism that weights each training sample based on the likelihood of them being normal samples or poisoned samples. The resulting weight vector can then be used in conjunction with any linear regression model that supports a weighted loss function (e.g., weighted linear least squares function). Therefore the proposed defense mechanism can be used to make learners such as ridge regression, LASSO regression, elastic-net regression, and neural network regression (NNR) resilient against poisoning attacks.

We first introduce a novel LID measure called \textit{Neighborhood LID ratio} (N-LID) that measures the local deviation of a particular sample's LID with respect to its $k$ nearest neighbors. Therefore N-LID can identify regions with similar LID values and samples that have significantly different LID values than their neighbors. We then show that N-LID values of poisoned and normal samples have two distinguishable distributions and introduce a baseline weighting mechanism based on the likelihood ratio of each data sample's N-LID (i.e., how many times more likely that the N-LID value is from the N-LID distribution of normal samples than the N-LID distribution of poisoned samples). Although the learners' capabilities are exaggerated, this paves the way for another N-LID based weighting scheme that assumes no knowledge of the attacker or the attacked samples. The latter defense has up to $76\%$ lower mean squared error (MSE) compared to an undefended ridge model, without the increased computational costs associated with prior works.

\begin{figure}[htp]
	\centering
	{\includegraphics[width=.6\columnwidth]{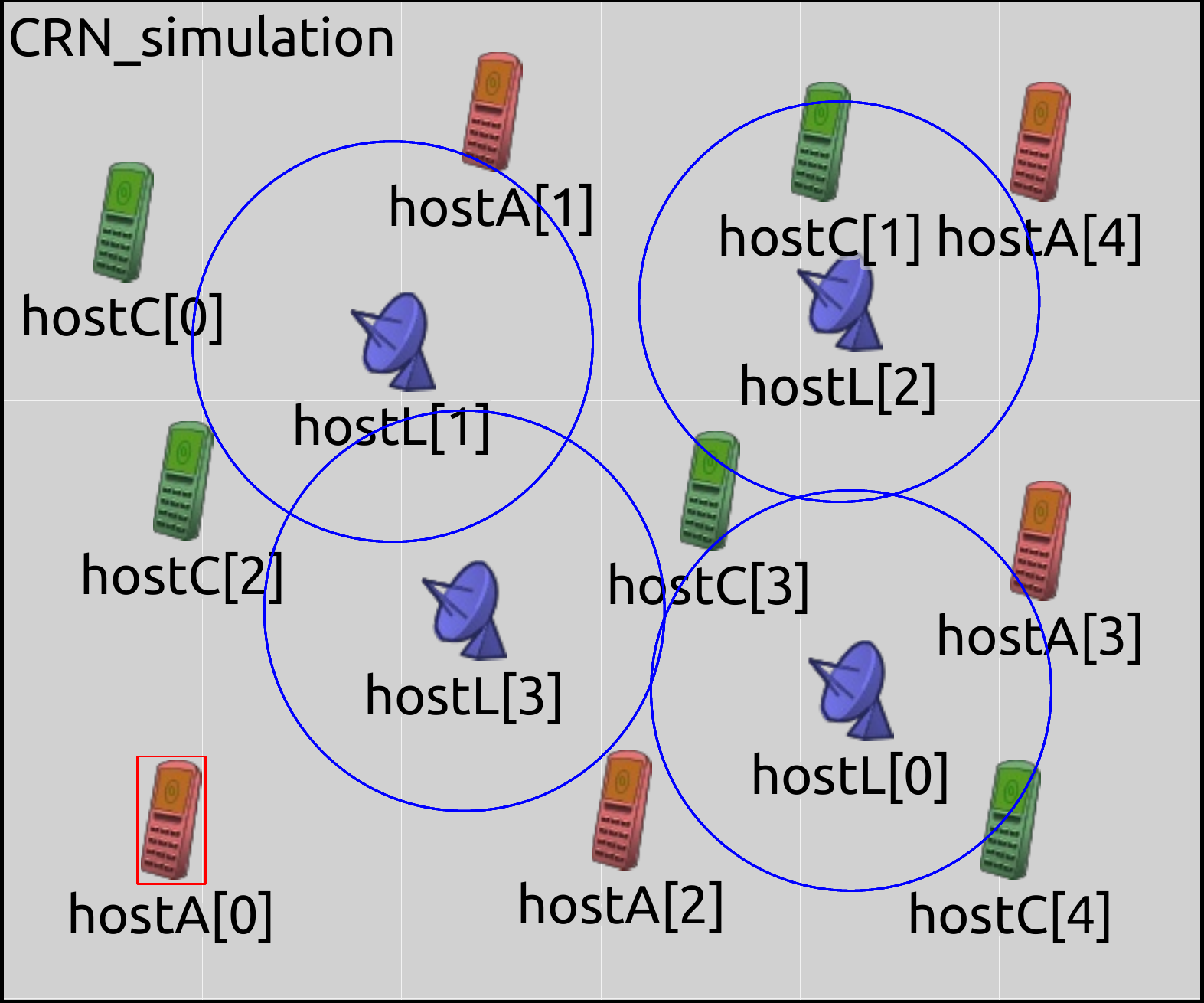}}
	\caption{A representation of the SDR listeners (blue), civilians (green) and rogue agents (red) in OMNET++ simulator.}
	\label{fig:omnet}
\end{figure}

As a real-world application of the proposed defense algorithm, we consider the following security application in the communications domain as a case study. Software-defined radios (SDRs) with considerable computing and networking capabilities can be utilized as an inexpensive scanner array for distributed detection of transmissions (Figure \ref{fig:omnet}). Based on the captured transmission signals, the transmission sources are assigned a risk value $\tau \in [0,1]$, based on the probability of them being malicious transmission sources (i.e., rogue agent). To clarify, background radio traffic (e.g., civilians) would have a $\tau$ value closer to zero, and a transmission source identified as a rogue agent would have a $\tau$ value closer to one. Due to the widespread presence of encryption methods, the risk association for transmission sources has to be inferred based on their statistical characteristics, without considering the transmitted information.

The task of assigning a risk value can be formulated as a regression problem where the learner creates an estimator based on data collected from known transmission sources. However, if the rogue agents deliberately try to mask their activities by altering their transmission patterns during the initial deployment of the SDR network (i.e., when the data is collected for training), they may force the learner to learn an estimator that is compromised (poisoning attack). Therefore, to prevent the estimator from being compromised, the learner uses a defense algorithm during the training phase. In Section \ref{sec:omnet_simulation}, we present details of the simulation setup followed by empirical results demonstrating the usefulness of our proposed defense.

The main \textbf{contributions} of this work are:
\begin{enumerate}
	\item A novel LID measure called \textit{Neighborhood LID ratio} that takes into account the LID values of a sample's $k$ nearest neighbors.
	\item An N-LID based defense mechanism that makes no assumptions regarding the attack, yet can be used in conjunction with several existing learners such as ridge, LASSO and NNR.
	\item Extensive numerical experiments that show N-LID weighted regression models provide significant resistance against poisoning attacks compared to state of the art alternatives.
\end{enumerate}

The remainder of the paper is organized as follows. Section \ref{sec:lit_review} provides details of previous literature relevant to this work. Section \ref{sec:problem_statement} formally defines the problem being addressed followed by Section \ref{sec:defense_method}, where we introduce the defense algorithms. We then describe the existing attacks against regression models in Section \ref{sec:reg_attack_model}. Section \ref{sec:regression_results} includes a detailed empirical analysis of the attacks and defenses on several real-world datasets followed by the results and discussion. The concluding remarks of Section \ref{sec:regression_conclusions} conclude the paper.

\section{Literature review}\label{sec:lit_review}
In this section, we briefly review the relevant previous work in the literature.

\subsection{Robust regression}
In robust statistics, robust regression is well studied to devise estimators that are not strongly affected by the presence of noise and outliers \cite{rousseeuw2005robust}. Huber \cite{huber1992robust} uses a modified loss function that optimizes the squared loss for relatively small errors and absolute loss for relatively large ones. As the influence on absolute loss by outliers is less compared to squared loss, this reduces their effect on the optimization problem, thereby resulting in a less distorted estimator. Mangasarian and Musicant \cite{mangasarian2000robust} solve the Huber M-estimator problem by reducing it to a quadratic program, thereby improving its performance. The Theil–Sen estimator, developed by Thiel \cite{thiel1950rank} and Sen \cite{sen1968estimates} uses the median of pairwise slopes as an estimator of the slope parameter of the regression model. As the median is a robust measure that is not influenced by outliers, the resulting regression model is considered to be robust as well.

Fischler and Bolles \cite{fischler1981random} introduced random sample consensus (RANSAC), which iteratively trains an estimator using a randomly sampled subset of the training data. At each iteration, the algorithm identifies data samples that do not fit the trained model by a predefined threshold as outliers. The remaining data samples (i.e., inliers) are considered as part of the consensus set. The process is repeated a fixed number of times, replacing the currently accepted model if a refined model with a larger consensus set is found. Huang et al. \cite{huang2015robust} decomposes the noisy data in to clean data and outliers by formulating an optimization problem. The estimator parameters are then computed using the clean data.
 
While these methods provide robustness guarantees against noise and outliers, we focus on malicious perturbations by sophisticated attackers that craft adversarial samples that are similar to normal data. Moreover, the aforementioned classical robust regression methods may fail in high dimensional settings as outliers might not be distinctive in that space due to high-dimensional noise \cite{xu2012outlier}.

\subsection{Adversarial regression}
The problem of learning under adversarial conditions has inspired a wide range of research from the machine learning community, see the work of Liu et al. \cite{liu2018survey} for a survey. Although adversarial learning in classification and anomaly detection has received a lot of attention \cite{tang2018adversarial,biggio2018wild,weerasinghe2019106985,karim2019adversarial}, adversarial regression remains relatively unexplored.

Most works in adversarial regression provide performance guarantees given several assumptions regarding the training data and noise distribution hold. The resilient algorithm by Chen et al. \cite{chen2013robust} assumes that the training data before adversarial perturbations, as well as the additive noise, are sub-Gaussian and that training features are independent. Feng et al. \cite{feng2014robust} provides a resilient algorithm for logistic regression under similar assumptions.

Xiao et al. \cite{xiao2015feature} examined how the parameters of the hyperplane in a linear regression model change under adversarial perturbations. The authors also introduced a novel threat model against linear regression models. Liu et al. \cite{liu2017robust} use robust PCA to transform the training data to a lower-dimensional subspace and follow an iterative trimmed optimization procedure where only the $n$ samples with the lowest residuals are used to train the model. Note that $n$ here is the number of normal samples in the training data set. Jagielski et al. \cite{jagielski2018manipulating} use a similar approach for the algorithm ``TRIM'', where trimmed optimization is performed in the input space itself. Therefore they do not assume a low-rank feature matrix as done in \cite{liu2017robust}. Both approaches, however, assume that the learner is aware of $n$ (or at least an upper bound for $n$), thereby reducing the practicality of the algorithms. 

Tong et al. \cite{tong2018adversarial} consider an attacker that perturbs data samples during the test phase to induce incorrect predictions. The authors use a single attacker, multiple learner framework modeled as a multi-learner Stackelberg game. Alfeld et al. \cite{alfeld2016data} propose an optimal attack algorithm against auto-regressive models for time series forecasting. In our work, we consider poisoning attacks against models that are used for interpolating instead of extrapolating. Nelson et al. \cite{nelson2008exploiting} originally introduced RONI (Reject On Negative Impact) as a defense for adversarial attacks against classification models. In RONI, the impact of each data sample on training is measured, and samples with notable negative impacts are excluded from the training process.

In summary, to the best of our knowledge, no linear regression algorithm has been proposed that reduces the effects of adversarial perturbations in training data through the use of LID. Most existing works on adversarial regression make strong assumptions regarding the attacker/attack, making them impractical, whereas, we propose a novel defense that makes no such assumptions.

\section{Problem statement}\label{sec:problem_statement}
We consider a linear regression learning problem in the presence of a malicious adversary. Define the labeled pristine training data (i.e., prior to adversarial perturbations) as $S:=(X,y)$, where $X\in\mathbb{R}^{n\times d}$ is the feature matrix and $y\in\mathbb{R}^{n\times 1}$ the corresponding response variable vector. We let $x_{i}\in\mathbb{R}^{d\times 1}$ denote the $i^{\text{th}}$ training sample, associated with a corresponding response variable $y_{i}\in\mathbb{R}$ from $y$. Note that each $y_{i}\in[0,1]$ for $i\in\{1,\ldots ,n\}$ unlike in classification problems where the labels take discrete categorical values. 

A linear regression model can be described by a linear function of the form $h_{\theta}(x_{i})=\omega^{T}x_{i}+b$ that predicts the response variable $\hat{y}_{i}$ for each $x_{i}\in X$. The function $h$ is parameterized by the vector $\theta=(\omega, b)\in \mathbb{R}^{d+1}$ consisting of the feature weights $\omega\in \mathbb{R}^{d}$ and the bias of the hyperplane $b\in \mathbb{R}$. The parameter vector $\theta$ is obtained by applying a learning algorithm, such as ridge regression (shown below), on $S$:
\begin{equation}\label{eq:clean_regression}
\begin{aligned}
&\theta^{*}=
& & \underset{\theta}{\argmin}
& & \frac{1}{n}\sum_{i=1}^{n}\big( y_{i}-h_{\theta}(x_{i})\big)^2 + \lambda \Vert \omega \Vert^2_2.\\
\end{aligned}
\end{equation}
Note that $\lambda$ is a regularization parameter.

The adversary's goal is to maximize the regression model's prediction error for unseen data samples (i.e., testing phase). To succeed in this, the adversary introduces a set of poisoned data samples $\bar{X}\in\mathbb{R}^{p\times d}$ and labels $\bar{y}\in\mathbb{R}^{p\times 1}$ into the pristine training set $S$, resulting in a contaminated training dataset $\tilde{S}:=(\tilde{X},\tilde{y})$ where $\tilde{X}=X\cup \bar{X}$ and $\tilde{y}=y\cup \bar{y}$. Applying a learning algorithm on the contaminated training data would result in a compromised regression model as follows,
\begin{equation}\label{eq:pois_regression}
\begin{aligned}
&\tilde{\theta}=
& & \underset{\theta}{\argmin}
& & \frac{1}{n+p}\sum_{i=1}^{n+p}\big( y_{i}-h_{\theta}(x_{i})\big)^2 + \lambda \Vert \omega \Vert^2_2,\\
\end{aligned}
\end{equation}
where $y_{i}\in \tilde{y}$ and $x_{i}\in \tilde{X}$. Through $\bar{X}$ and $\bar{y}$, the adversary forces the learner to obtain a parameter vector $\tilde{\theta}$ that is significantly different from $\theta^*$, which it would have obtained had the training data been unaltered.

To address this problem, we propose an LID based weighting scheme that can be incorporated into learning algorithms that optimize quadratic loss functions such as the Mean Squared Error in Equation (\ref{eq:pois_regression}). In \textit{weighted least squares (WLS)}, a weight $\beta_{i}\in[0,1]$ is assigned for each sample in order to discriminate and vary their influence on the optimization problem. A small $\beta_{i}$ value would allow for a large residual value, and the effect of the sample would be de-emphasized. Conversely, a large $\beta_{i}$ value would emphasize that particular sample's effect. The weighted regression learning problem is formulated as the following convex optimization problem:
\begin{equation}\label{eq:weighted_regression}
\begin{aligned}
&\hat{\theta}=
& & \underset{\theta}{\argmin}
& & \frac{1}{n+p}\sum_{i=1}^{n+p}\beta_{i}\big( y_{i}-h_{\theta}(x_{i})\big)^2 + \lambda \Vert \omega \Vert^2_2.\\
\end{aligned}
\end{equation}
By carefully selecting a weight vector $\beta$, the learner can minimize the effects of the adversarial perturbations and recover the correct estimator $\hat{\theta}\approx\theta^{*}$. 

\section{Defense algorithms}\label{sec:defense_method}
First, we briefly introduce the theory of LID for assessing the dimensionality of data subspaces, then, we present our novel LID based mechanism to obtain the weight vector $\beta$ and show how it can easily be incorporated into existing regression algorithms.

Expansion models of dimensionality have previously been successfully employed in a wide range of applications, such as manifold learning, dimension reduction, similarity search, and anomaly detection \cite{LID1_Houle,amsaleg2015estimating}. \emph{In this paper, we use LID to characterize the intrinsic dimensionality of regions where attacked samples lie, and create a weighting mechanism that de-emphasizes the effect of samples that have a high likelihood of being adversarial examples.}

\subsection{Theory of Local Intrinsic Dimensionality.}\label{sec:lid_theory}
In the theory of intrinsic dimensionality, classical expansion models measure the rate of growth in the number of data samples encountered as the distance from the sample of interest increases \cite{LID1_Houle}. As an example, in Euclidean space, the volume of an m-dimensional ball grows proportionally to $r^m$, when its size is scaled by a factor of $r$. From this rate of change of volume w.r.t distance, the expansion dimension $m$ can be deduced as:
\begin{equation}
\frac{V_{2}}{V_{1}}=
\left(\frac{r_{2}}{r_{1}}\right)^m\Rightarrow m=\frac{\ln\left(V_{2}/V_{1}\right)}{\ln\left(r_2/r_1\right)}.\\
\end{equation}

Transferring the concept of expansion dimension to the statistical setting of continuous distance distributions leads to the formal definition of LID. By substituting the cumulative distance for volume, LID provides measures of the intrinsic dimensionality of the underlying data subspace. Refer to the work of Houle \cite{LID1_Houle} for more details concerning the theory of LID. The formal definition of LID is given below \cite{LID1_Houle}.\\

\begin{figure}[t]
	\centering
	\begin{subfigure}[t]{0.12\textwidth} 
		\includegraphics[width=\textwidth]{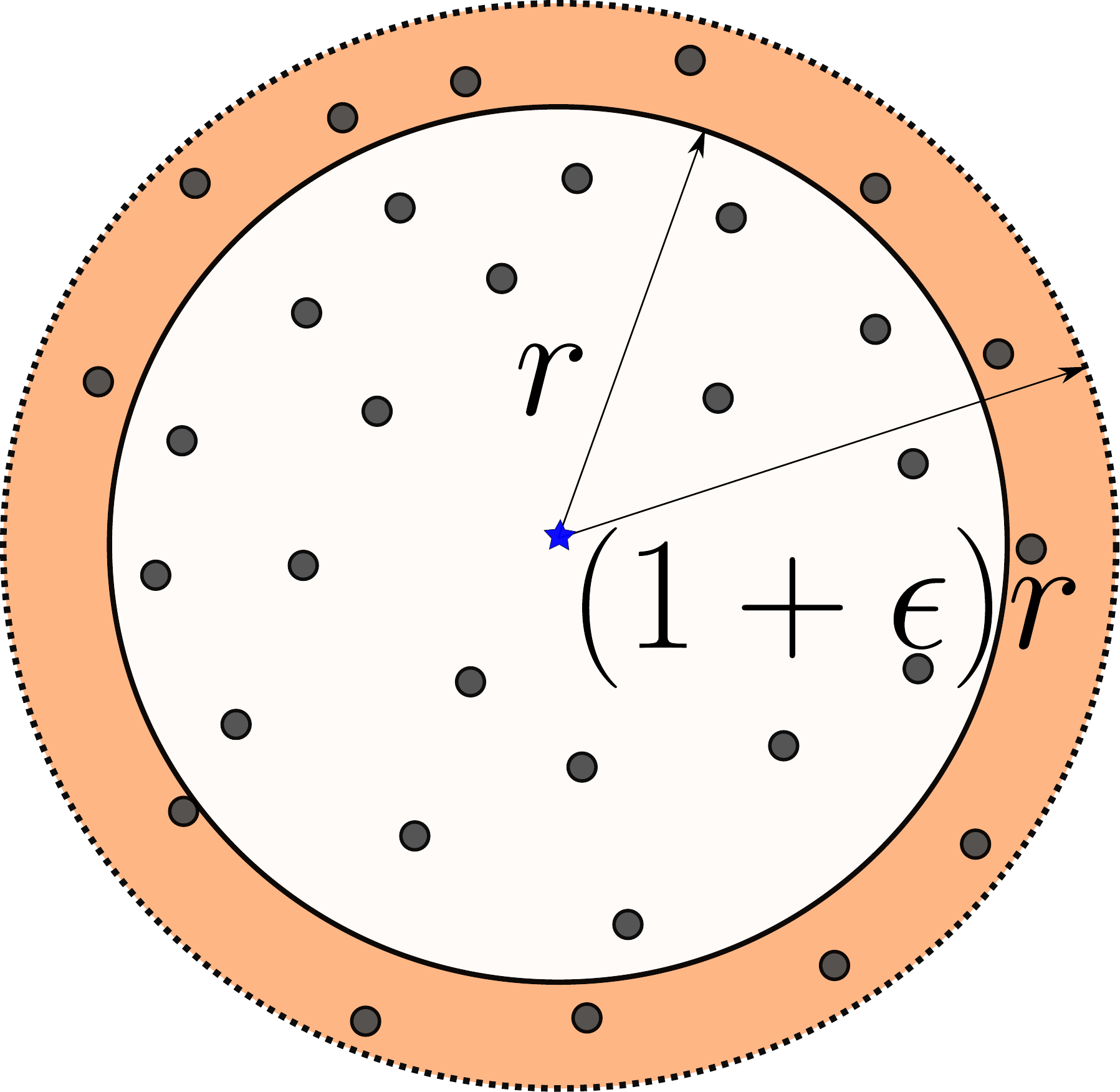}
	\end{subfigure}
	\begin{subfigure}[t]{0.23\textwidth} 
		\includegraphics[width=\textwidth]{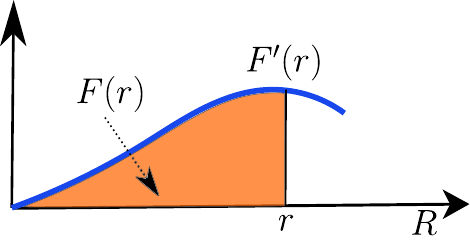}
	\end{subfigure}
	\caption{The left figure shows the distribution of data points around the sample of interest (blue star) in a $d$-dimensional Euclidean space. The right figure shows the corresponding distribution of distance.}
	\label{fig:lid_description}
\end{figure}

\noindent\textbf{Definition 1} (Local Intrinsic Dimensionality).\\
	\textit{Given a data sample $x\in X$, let $R > 0$ be a random variable denoting the distance from $x$ to other data samples. If the cumulative distribution function $F(r)$ of $R$ is positive and continuously differentiable at distance $r > 0$, the LID of $x$ at distance $r$ is given by:}
	\begin{equation}\label{eq:lid} 
	\text{LID}_{F}(r)\triangleq \Lim{\epsilon \rightarrow 0}\frac{\text{ln}\big(F((1+\epsilon)\cdot r)/F(r)\big)}{\text{ln}(1+\epsilon)}=\frac{r\cdot F'(r)}{F(r)},
	\end{equation}
\textit{whenever the limit exists.}

The last equality of Equation (\ref{eq:lid}) follows by applying L'H\^{o}pital's rule to the limits \cite{LID1_Houle}. The local intrinsic dimension at $x$ is in turn defined as the limit when the radius $r$ tends to zero:
\begin{equation}\label{eq:lid_2}
\text{LID}_{F}=\Lim{r \rightarrow 0}\text{LID}_{F}(r).
\end{equation}

$\text{LID}_{F}$ describes the relative rate at which its cumulative distribution function $F(r)$ increases as the distance $r$ increases from $0$ (Figure \ref{fig:lid_description}). In the ideal case where the data in the vicinity of $x$ is distributed uniformly within a subspace, $\text{LID}_{F}$ equals the dimension of the subspace; however, in practice these distributions are not ideal, the manifold model of data does not perfectly apply, and $\text{LID}_{F}$ is not an integer \cite{LID_sarah_ICLR}. Nevertheless, the local intrinsic dimensionality is an indicator of the dimension of the subspace containing $x$ that would best fit the data distribution in the vicinity of $x$.\newline

\noindent\textbf{Definition 2} (Estimation of LID).\\
	\textit{Given a reference sample $x \sim \mathcal{P}$, where $\mathcal{P}$ represents the data distribution, the Maximum Likelihood Estimator of the LID at $x$ is defined as follows \cite{amsaleg2015estimating}:}
	\begin{equation}\label{eq:lid_estimation}
	\widehat{\text{LID}}(x)=-\Bigg(\frac{1}{k}\sum_{i=1}^{k}\text{log}\frac{r_{i}(x)}{r_{\text{max}}(x)}\Bigg)^{-1}.
	\end{equation}
	\textit{Here, $r_{i}(x)$ denotes the distance between $x$ and its $i$-th nearest neighbor within a sample of $k$ points drawn from $\mathcal{P}$, and $r_{\text{max}}(x)$ is the maximum of the neighbor distances.}

The above estimation assumes that samples are drawn from a tight neighborhood, in line with its development from Extreme Value Theory. In practice, the sample set is drawn uniformly from the available training data (omitting $x$ itself), which itself is presumed to have been randomly drawn from $\mathcal{P}$. Note that the LID defined in Equation (\ref{eq:lid_2}) is the theoretical calculation, and that $\widehat{\text{LID}}$ defined in Equation (\ref{eq:lid_estimation}) is its estimate.

\subsection{Neighborhood LID Ratio.}\label{sec:lid_poisoned}
We now introduce a novel LID based measure called neighborhood LID ratio (N-LID) and discuss the intuition behind using N-LID to identify adversarial samples during training. By perturbing the feature vector, the adversary moves a training sample away from the distribution of normal samples. Computing LID estimates with respect to its neighborhood would then reveal an anomalous distribution of the local distance to these neighbors. Furthermore, as $\widehat{\text{LID}}(x)$ is an indicator of the dimension of the subspace that contains $x$, by comparing the LID estimate of a data sample to the LID estimates of its nearest neighbors, we can identify regions that have a similar lower-dimensional subspace. More importantly, any samples that have a substantially different lower-dimensional subspace compared to its neighbors can be identified as poisoned samples.

Considering this aspect of poisoning attacks, we propose a novel LID ratio measure that has similar properties as the local outlier factor (LOF) algorithm introduced by Breunig et al. \cite{breunig2000lof} as follows:\\

\noindent\textbf{Definition 3} (Neighborhood LID ratio).\\
\textit{Given a data sample $x \in X$, the neighborhood LID ratio of $x$ is defined as:}
\begin{equation}\label{eq:neighborhood_lid_ratio}
\text{N-LID}(x)=\dfrac{\frac{1}{k}\sum_{i=1}^{k}	\widehat{\text{LID}}(x_{i})}{\widehat{\text{LID}}(x)}.
\end{equation}

\begin{figure*}[t]
	\centering
	\begin{subfigure}[t]{0.4\textwidth} 
		\includegraphics[width=\textwidth]{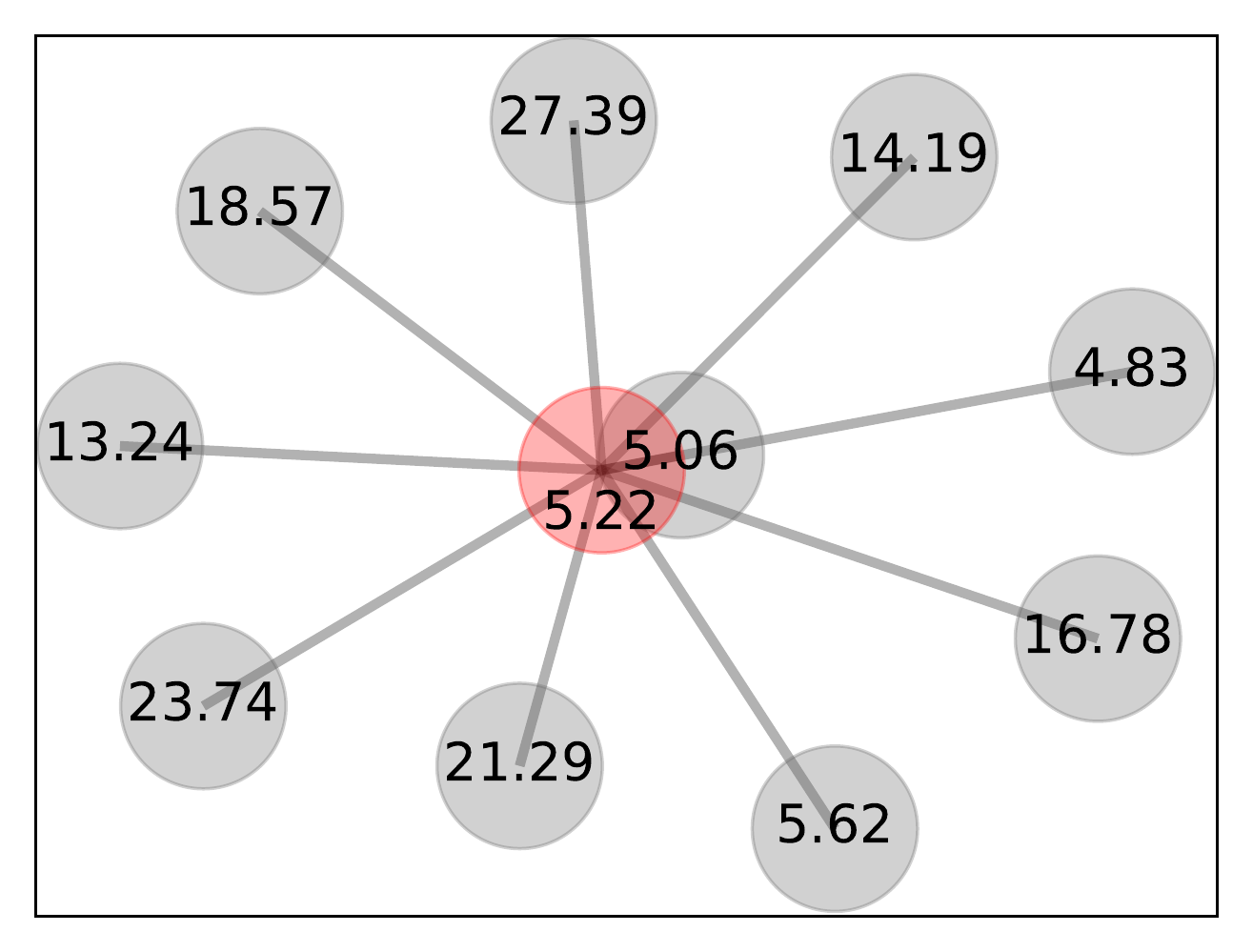}
		\caption{N-LID calculation of a poisoned sample}
		\label{fig:nlid_graph}
	\end{subfigure}
	\begin{subfigure}[t]{0.4\textwidth} 
		\includegraphics[width=\textwidth]{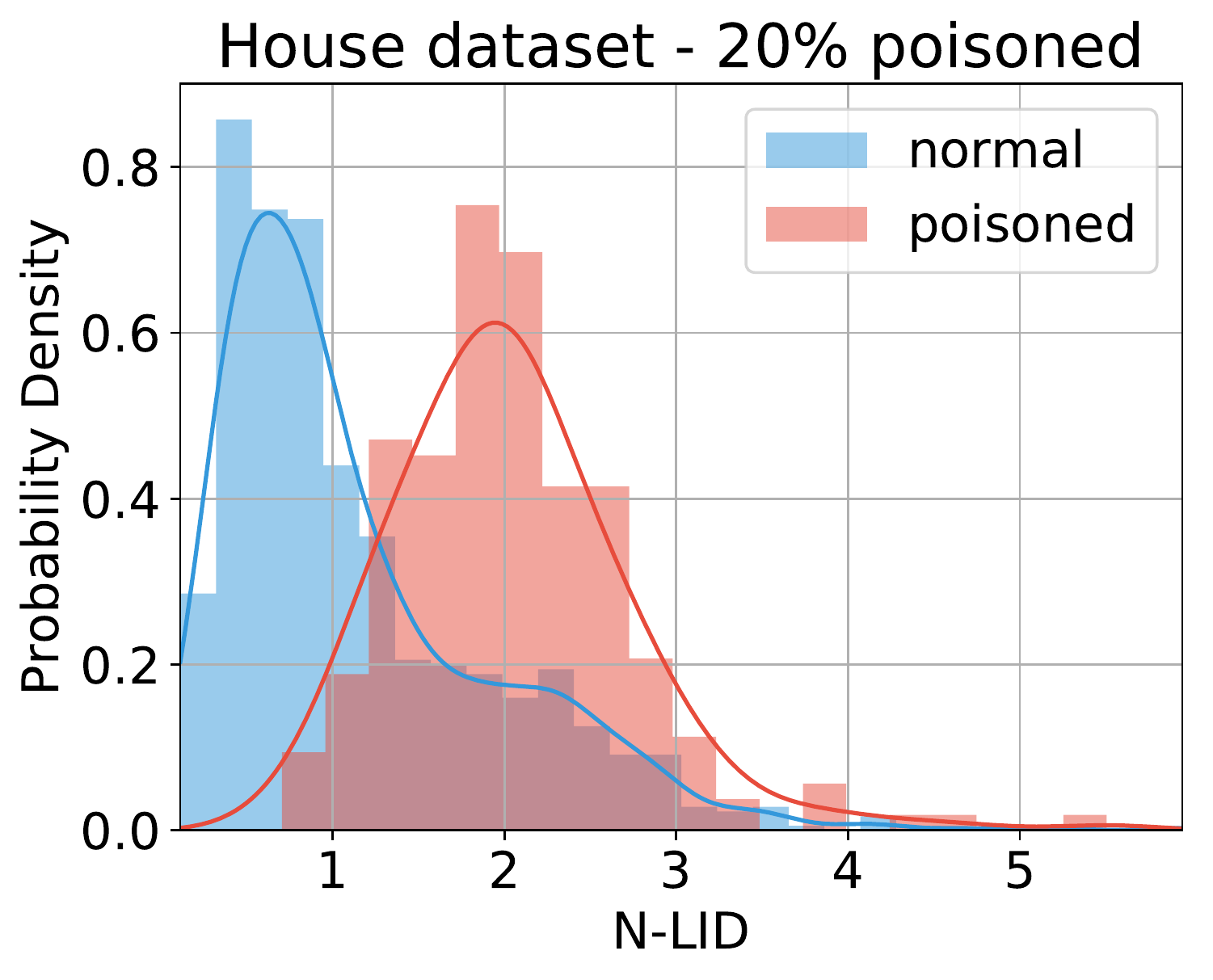}
		\caption{N-LID distributions in House dataset.}
		\label{fig:pdf}
	\end{subfigure}
	\caption{The left figure shows the LID values of a poisoned sample (red) and its $k$ nearest neighbors. The poisoned sample has an N-LID of $2.89$. The right figure shows the N-LID distributions of the normal samples and poisoned samples when 20\% of the data is poisoned.}
	\label{fig:nlid_distr}
\end{figure*}

Here, $\widehat{\text{LID}}(x_{i})$ denotes the LID estimate of the $i$-th nearest neighbor from the $k$ nearest neighbors of $x$. Figure \ref{fig:nlid_graph} shows the LID estimates of a poisoned sample and its $k$ nearest neighbors. As the poisoned sample's LID estimate is substantially different from its neighbors, the N-LID value calculated using Equation(\ref{eq:neighborhood_lid_ratio}) highlights it as an outlier. As Figure \ref{fig:pdf} shows, N-LID is a powerful metric that can give two distinguishable distributions for poisoned and normal samples. Although ideally, we like to see no overlap between the two distributions, in real-world datasets, we see some degree of overlap.

\subsection{Baseline Defense.}\label{sec:lr_defense}
We now describe a baseline weighting scheme calculated using the N-LID distributions of poisoned and normal samples. Define $p_{n}$ and $p_{a}$ as the probability density functions of the N-LID values of normal samples and poisoned samples. For a given data sample $x_{i}$ with the N-LID estimate $\text{N-LID}(x_{i})$, we define two possible hypotheses, $H_{n}$ and $H_{a}$ as
\begin{equation}\label{eq:hypothesis}
\begin{aligned}
& H_{n}:\text{N-LID}(x_{i})\sim p_{n},\\
& H_{a}:\text{N-LID}(x_{i})\sim p_{a},
\end{aligned}
\end{equation}
where the notation ``$\text{N-LID}(x_{i})\sim p$'' denotes the condition ``$\text{N-LID}(x_{i})$ is from the distribution $p$''. To clarify, $H_{n}$ is the hypothesis that $\text{N-LID}(x_{i})$ is from the N-LID distribution of normal samples and $H_{a}$ is the hypothesis that $\text{N-LID}(x_{i})$ is from the N-LID distribution of poisoned samples. 

The \textit{likelihood ratio} (LR) is usually used in statistics to compare the goodness of fit of two statistical models. We define the likelihood ratio of a data sample $x_{i}$ with the N-LID estimate $\text{N-LID}(x_{i})$ as
\begin{equation}\label{eq:likelihood_ratio}
\Lambda\big(\text{N-LID}(x_{i})\big)={p_{n}\big(\text{N-LID}(x_{i})\big)}/{p_{a}\big(\text{N-LID}(x_{i})\big)},
\end{equation}
where $p(\text{N-LID}(x_{i}))$ denotes the probability of the N-LID of sample $x_{i}$ w.r.t the probability distribution $p$. To clarify, $\Lambda(\text{N-LID}(x_{i}))$ expresses how many times more likely it is that $\text{N-LID}(x_{i})$ is under the N-LID distribution of normal samples than the N-LID distribution of poisoned samples. 

As there is a high possibility for $p_{n}$ and $p_{a}$ to have an overlapping region (Figure \ref{fig:pdf}), there is a risk of emphasizing the importance of (giving a higher weight to) poisoned samples. To mitigate that risk, we only de-emphasize samples that are suspected to be poisoned (i.e., low LR values). Therefore, we transform the LR values such that $\Lambda(\text{N-LID}(x))\in[0,1]$. The upper bound is set to $1$ to prevent overemphasizing samples. 

Subsequently, we fit a hyperbolic tangent function to the transformed LR values ($z$) in the form of $0.5(1 - \tanh(az - b))$ and obtain suitable values for the parameters $a$ and $b$. The scalar value of $0.5$ maintains the scale and vertical position of the function between $0$ and $1$. By fitting a smooth function to the LR values, we remove the effect of noise and enable the calculation of weights for future training samples. Finally, we use the N-LID value of each $x_{i}$ and use Equation (\ref{eq:function_fitting}) to obtain its corresponding weight  $\beta_{i}$.  
\begin{equation}\label{eq:function_fitting}
\beta_{i}=0.5\big(1 - \tanh(a\text{N-LID}(x_{i}) - b)\big). 
\end{equation}

\subsection{Attack Unaware Defense.}\label{sec:unaware_defense}
The LR based weighting scheme described above assigns weights to training samples based on the probabilities of their N-LID values. This approach requires the learner to be aware of the probability density functions (PDFs) of normal and poisoned samples. The two distributions can be obtained by simulating an attack and deliberately altering a subset of data during training, by assuming the distributions based on domain knowledge or prior experience or by having an expert identify attacked and non-attacked samples in a subset of the dataset. However, we see that the N-LID measure in Equation (\ref{eq:neighborhood_lid_ratio}) results in larger values for poisoned samples compared to normal samples (Figure \ref{fig:pdf}). In general, therefore, it seems that a weighting scheme that assigns small weights to large N-LID values and large weights to small N-LID values would lead to an estimator that is less affected by the poisoned samples present in training data.

\begin{figure}[h]
	\centering
	\begin{subfigure}{0.5\textwidth} 
		\includegraphics[width=\textwidth]{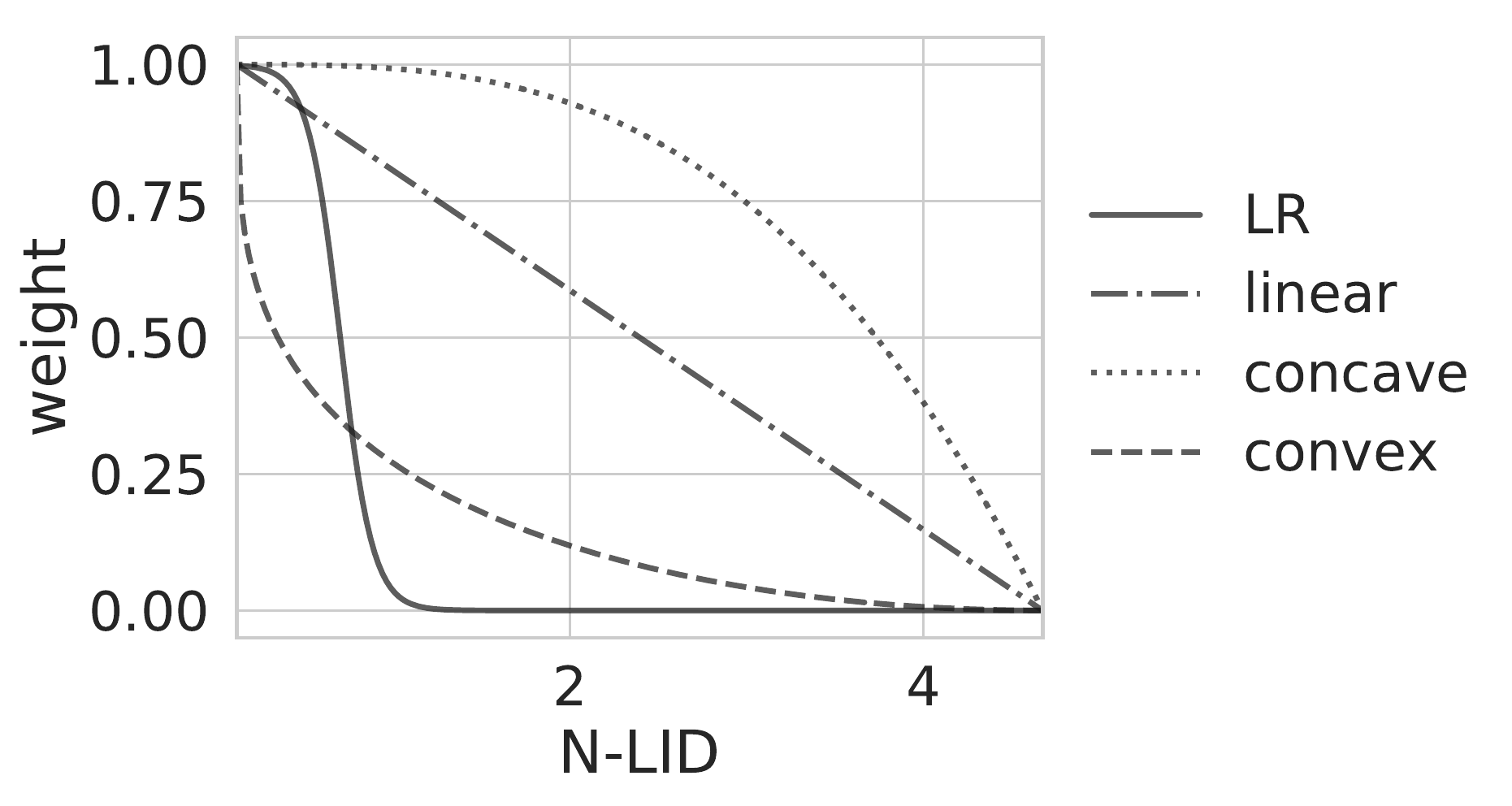}
	\end{subfigure}
	\caption{The N-LID based weighting schemes.}
	\label{fig:weight_functions}
\end{figure}

Considering this aspect of N-LID, we choose three weighting mechanisms that assign $\beta=1$ for the sample with the smallest N-LID and $\beta=0$ for the sample with the largest N-LID as shown in Figure \ref{fig:weight_functions}. Note that there is a trade-off between preserving normal samples, and preventing poisoned samples from influencing the learner. The concave weight function attempts to preserve normal samples, which causes poisoned samples to affect the learner as well. In contrast, the convex weight function reduces the impact of poisoned samples while reducing the influence of many normal samples in the process. The linear weight function assigns weights for intermediate N-LID values linearly without considering the trade-off. 

We obtain the weight functions by first scaling the N-LID values of the training samples to $[0,1]$. Take $v_{i}$ as the scaled N-LID value of sample $x_{i}$. Then, for the linear weight function, we take $\beta_{i}=1/v_{i}$ as the weight. For the concave weight function we use $\beta_{i}=1 - v_{i}^2$ and for the convex weight function we use $\beta_{i}=1 - (2v_{i} - v_{i}^2)^{0.5}$. We choose these two arbitrary functions as they possess the trade-off mentioned in the paragraph above. Any other functions that have similar characteristics can also be used to obtain the weights.

The main advantage of our proposed defense is that it is decoupled from the learner; it can be used in conjunction with any learner that allows the loss function to be weighted. In Section \ref{sec:regression_results}, we demonstrate the effectiveness of our proposed defense by incorporating it into a ridge regression model and an NNR model. 

The high level procedures used to obtain uninfluenced ridge regression models under adversarial conditions are formalized in Algorithms \ref{algo:lid_regression_lr} and \ref{algo:lid_regression} for the baseline defense (N-LID LR) and the convex weight function based defense (N-LID CVX).

\begin{algorithm}[t]
	{\small
		\caption{N-LID LR Defense Algorithm}\label{algo:lid_regression_lr}
		\begin{algorithmic}[1]
			\State \textbf{input} $\tilde{S}=(\tilde{X},\tilde{y})$\Comment{contaminated training data} 
			\State \textbf{output} uninfluenced ridge regression model
			\State initialize $\text{N-LID}_{a},\text{N-LID}_{n},\Lambda,\beta=[~]$
			\State $\text{N-LID}_{a} \gets{{\text{n-lid-calculation}}(\tilde{S})}$\Comment{calculate n-lid values of poisoned samples} 
			\State $\text{N-LID}_{n} \gets{{\text{n-lid-calculation}}(\tilde{S})}$\Comment{calculate n-lid values of normal samples}  
			\State $\Lambda \gets\text{calculate likelihood ratio for each N-LID value}$
			\State fit function $g(z)=0.5(1 - \tanh(az - b))$ to $\Lambda$ to find $a$ and $b$			 
			\State $\beta \gets g(\text{N-LID}(x))$\Comment{obtain weight $\beta_{i}$ of each $x_{i}\in X$ using the N-LID of $x_{i}$}
			\State N-LID LR ridge = $\text{train-weighted-ridge-model}(\tilde{X},\tilde{y},\beta)$
		\end{algorithmic}}
\end{algorithm}

\begin{algorithm}[t]
	{\small
		\caption{N-LID CVX Defense Algorithm}\label{algo:lid_regression}
		\begin{algorithmic}[1]
			\State \textbf{input} $\tilde{S}=(\tilde{X},\tilde{y})$\Comment{contaminated training data} 
			\State \textbf{output} uninfluenced ridge regression model
			\State initialize $\text{N-LID}, v,\beta=[~]$
			\State $\text{N-LID} \gets{{\text{n-lid-calculation}}(\tilde{S})}$\Comment{calculate n-lid values of all samples} 
			\State $v \gets{{\text{scaler}}(\text{N-LID})}$\Comment{scale the N-LID values to $[0,1]$} 
			\State $\beta \gets 1 - (2v - v^2)^{0.5}$\Comment{obtain weight $\beta_{i}$ of each $x_{i}$ using the chosen convex function} 
			\State N-LID CVX ridge = $\text{train-weighted-ridge-model}(\tilde{X},\tilde{y},\beta)$
		\end{algorithmic}}
\end{algorithm}

\section{Threat models}\label{sec:reg_attack_model}
In this work, we assume a learner that does not have access to a collection of pristine data samples. The training data it uses may or may not be manipulated by an adversary. The learner chooses its defense algorithm and regression model, without knowing details of the attack employed by the adversary.

The attacks being considered are \textit{white-box} attacks, with the attacker knowing the pristine training data and any hyper-parameters of the learning algorithm. While these assumptions exaggerate the capabilities of a real-world attacker, it allows us to test the performance of defense algorithms under a worst-case scenario. Moreover, relying on secrecy for security is considered as a poor practice when designing defense algorithms \cite{biggio2014security}. 

We employ the attack algorithm introduced by Jagielski et al. \cite{jagielski2018manipulating} against linear regression models, which is an adaptation of the attack algorithm by Xiao et al. \cite{xiao2015feature}. The latter attack algorithm is used in prior works on adversarial regression as the threat model \cite{liu2017robust}. While giving a brief description of the attack algorithm, we refer the readers to \cite{jagielski2018manipulating} for more details.

The underlying idea is to move selected data samples along the direction that maximally increases the MSE of the estimator on a pristine validation set. Unlike in classification problems, the attacker can alter the response variable $y$ as well. We use similar values used by Jagielski et al. \cite{jagielski2018manipulating} for the hyper-parameters that control the gradient step and convergence of the iterative search procedure of the attack algorithm. The line search learning rate ($\eta$), which controls the step size taken in the direction of the gradient, is selected from a set of predefined values by evaluating the MSE values on a validation set.

The amount of poisoning is controlled by the \textit{poisoning rate}, defined as $p/(p+n)$, where $n$ is the number of pristine samples, and $p$ is the number of poisoned samples. The attacker is allowed to poison up to $20\%$ of the training data in line with prior works in the literature \cite{jagielski2018manipulating,xiao2015feature}. 

The Opt attack algorithm selects the initial set of data points to poison randomly from the training set. The corresponding response variables $\bar y$ are initialized in one of two methods: (i) $\bar y = 1-y$ and (ii) $\bar y = \text{round}(1-y)$. The first approach, known as Inverse Flipping (IFlip), results in a less intense attack compared to the latter, known as \textit{Boundary Flipping} (BFlip), which pushes the response values to the extreme edges. In our experiments, we test against attacks that use both initialization techniques.

We also considered the optimization based attack by Tong et al. \cite{tong2018adversarial}, but it was unable to exact a noticeable increase in MSE even at a poisoning rate of $20\%$. Therefore we do not present experimental results for it. 

\section{Experimental results and discussion}\label{sec:regression_results}
In this section, we describe the datasets used and other procedures of the experimental setup. We extensively investigate how the performance of N-LID weighted regression models holds against an increasing fraction of poisoned training data. Our code and datasets are available at \url{https://github.com/sandamal/lid-regression}.

\subsection{Case study: risk assignment to suspicious transmissions}\label{sec:omnet_simulation}
We provide here a brief description of the simulations that were conducted to obtain the data for the case study that we are considering in this paper. Refer to the work of Weerasinghe et al. \cite{weerasinghe2018detection} for additional information regarding the security application in SDR networks.

\textbf{Network simulation and data collection:}
We use the INET framework for OMNeT++ \cite{Varga:2008:OOS:1416222.1416290} as the network simulator, considering signal attenuation, signal interference, background noise, and limited radio ranges in order to conduct realistic simulations. We place the nodes (civilians, rogue agents, and listeners) randomly within the given confined area. The simulator allows control of the frequencies and bit rates of the transmitter radios, their communication ranges, message sending intervals, message lengths, the sensitivity of the receivers, minimum energy detection of receivers, among other parameters. It is assumed that all the communications are encrypted; therefore, the listeners are unable to access the content of the captured transmissions. We obtain the duration of the reception, message length, inter-arrival time (IAT), carrier frequency, bandwidth, and bitrate as features using the data captured by the listeners.

Considering the data received by the three closest listeners (using the power of the received signal) of each transmission source, the duration, message length, and IAT of the messages received by each listener are averaged every five minutes, which results in $108$ ($12\times 3 \times 3$) features in total. Adding the latter three parameters (fixed for each transmission source) gives the full feature vector of $111$ features. The response variables (i.e., the risk of being a rogue agent) are assigned based on the positions of the data samples in the feature space.

\subsection{Benchmark datasets.}
We use the following combination of relatively high dimensional and low dimensional regression datasets in our experimental analysis. In particular, The House and Loan datasets are used as benchmark datasets in the literature \cite{jagielski2018manipulating}. Table \ref{tab:dataset_description} provides the training and test set sizes, the regularization parameter $\lambda$ of the ridge regression model, and the line search learning rates $\eta$ of the attacks considered for each of the datasets.

\textbf{House dataset:} The dataset uses features such as year built, floor area, number of rooms, and neighborhood to predict the sale prices of houses. There are $275$ features in total, including the response variable. All features are normalized into $[0,1]$. We fix the training set size to $840$, and the test set size to $280$. 

\textbf{Loan dataset:} The dataset contains information regarding loans made on \textit{LendingClub}. Each data sample contains $89$ features, including loan amount, term, the purpose of the loan, and borrower's attributes with the interest rate as the response variable. We normalize the features into $[0,1]$ and fix the training set and test set size as above. 

\textbf{Grid dataset:} Arzamasov et al. \cite{arzamasov2018towards} simulated a four-node star electrical grid with centralized production to predict its stability, w.r.t the behavior of participants of the grid (i.e., generator and consumers). The system is described with differential equations, with their variables given as features of the dataset. In our work, we use the real component of the characteristic equation root as the response variable for the regression problem.

\textbf{Machine dataset:} The dataset by Kibler et al. \cite{kibler1989instance} is used to estimate the relative performance of computer hardware based on attributes such as machine cycle time, memory, cache size, and the number of channels.
 
{ 
\setlength{\tabcolsep}{4.5pt}
\begin{table}[h]
	\singlespacing
	\centering
	\caption{Description of the datasets.}
	\label{tab:dataset_description}
	\begin{tabular}{lrrrrrr}
		\toprule
Dataset & \makecell{\# of\\features} & \makecell{Training\\size} & \makecell{Test\\size} & $\lambda$ & \makecell{$\eta$ of\\BFlip} & \makecell{$\eta$ of\\IFlip} \\\midrule
House   & 274                & 840                                   & 280                               & 0.0373                        & 0.01                                & 0.01                                \\
Loan    & 88                 & 840                                   & 280                               & 0.0273                        & 0.05                                & 0.05                                \\
OMNeT & 111                  & 502                                   & 125                                & 0.0002                        & 0.01                                & 0.01                               \\
Grid    & 12                 & 840                                   & 280                               & 0.0173                        & 0.01                                & 0.30                                \\
Machine & 6                  & 125                                   & 42                                & 0.0223                        & 0.03                                & 0.03                               \\
		\bottomrule
	\end{tabular}
\end{table}
}

\subsection{Experimental setup.} For each learning algorithm, we find its hyperparameters using five-fold cross-validation on a pristine dataset. For LID calculations we tune $k$ over $[10,100)$ as previously done by Ma et al. \cite{LID_sarah_ICLR}. Kernel density estimation for LR calculation was performed using a Gaussian kernel with the bandwidth found using a cross-validation approach to avoid overfitting. The effects of kernel density estimation are neutralized by the smooth function fitting (Equation (\ref{eq:function_fitting})); therefore, the algorithm is robust to these choices.

For each dataset, we create five cross-validation sets. We present the average MSE of the cross-validation sets for each learner when the poisoning rate increases from $0$ to $20$. While increasing the poisoning rate over $20\%$ is feasible, it is unrealistic to expect a learner to withstand such amounts of poisoned data, and it also assumes a powerful attacker with significant influence over the training data. As the existence of such an attacker is unlikely, we limit the poisoning rate to $20\%$, similar to prior works in the literature. 

\subsection{Results and discussion.}

\begin{figure*}[h!]
	\centering
	\begin{subfigure}{\textwidth} 
		\includegraphics[width=\textwidth]{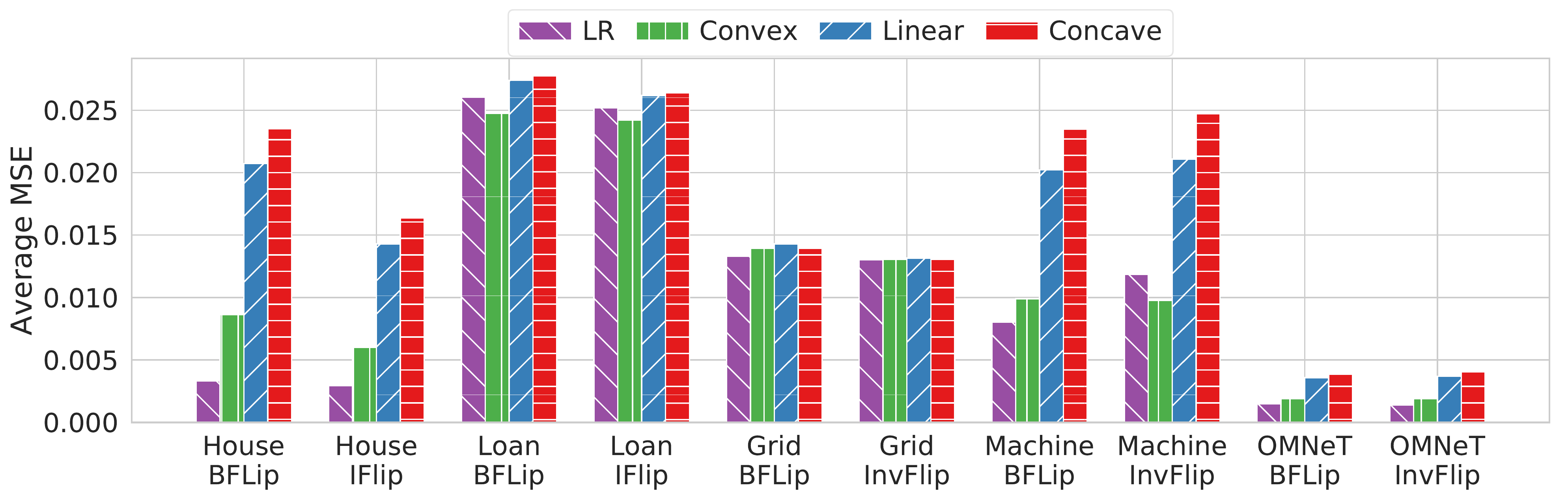}
	\end{subfigure}
	\caption{The average MSE for each weighting scheme under the two attack configurations (BFlip and IFlip) for the benchmark datasets and the OMNeT case study.}
	\label{fig:weight_functions_results}
\end{figure*}

\subsubsection{Attacker unaware defense}
Figure \ref{fig:weight_functions_results} compares the average MSE for the three attacker unaware weighting schemes and the LR based weighting scheme that we introduced in Section \ref{sec:unaware_defense}. Of the three attacker unaware weighting schemes, we see that the convex weight function has $58\%$ lower average MSE in House, $9\%$ in Loan, $51\%$ in Machine, and $47\%$ in OMNeT. This is not surprising as the convex weight function is the closest to the ideal LR based weight function. The result suggests that it is favorable to the learner to reduce the weights of suspected adversarial samples even if there is an increased risk of de-emphasizing normal samples as well. This is to be expected with linear regression data where there is high redundancy; losing a portion of normal data from training would not significantly impact the performance of the estimator. In future studies, it might be possible to investigate the performance of the three weighting functions in situations where data redundancy is low, such as non-linear regression problems. In the following sections, the term \textit{N-LID CVX} refers to a ridge regression model with the convex weight function based defense.

\begin{figure*}[t]
	\centering
	\begin{subfigure}[t]{0.4\textwidth} 
		\includegraphics[width=\textwidth]{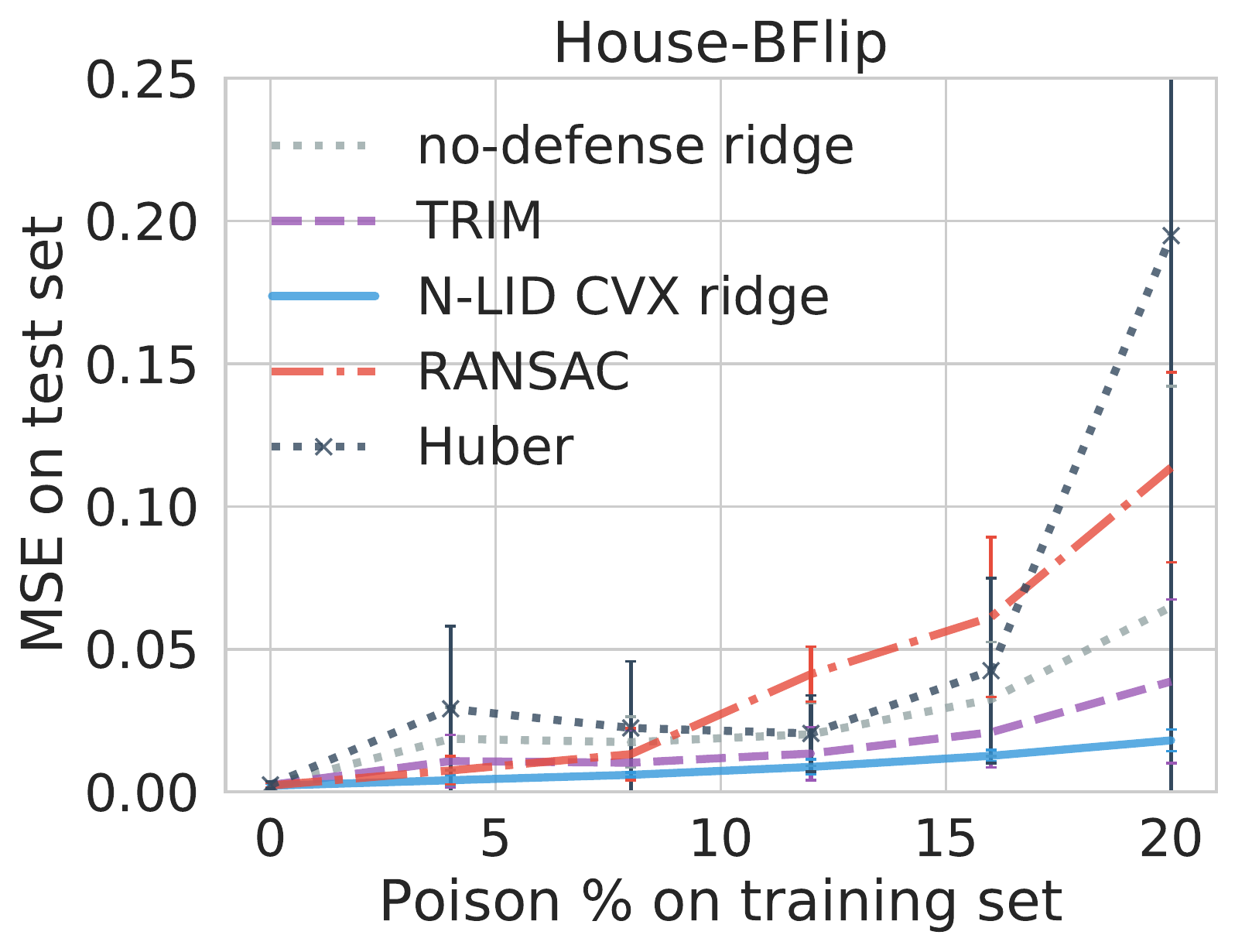}
		\caption{MSE on the House dataset.} 
	\end{subfigure}
	\begin{subfigure}[t]{0.4\textwidth} 
		\includegraphics[width=\textwidth]{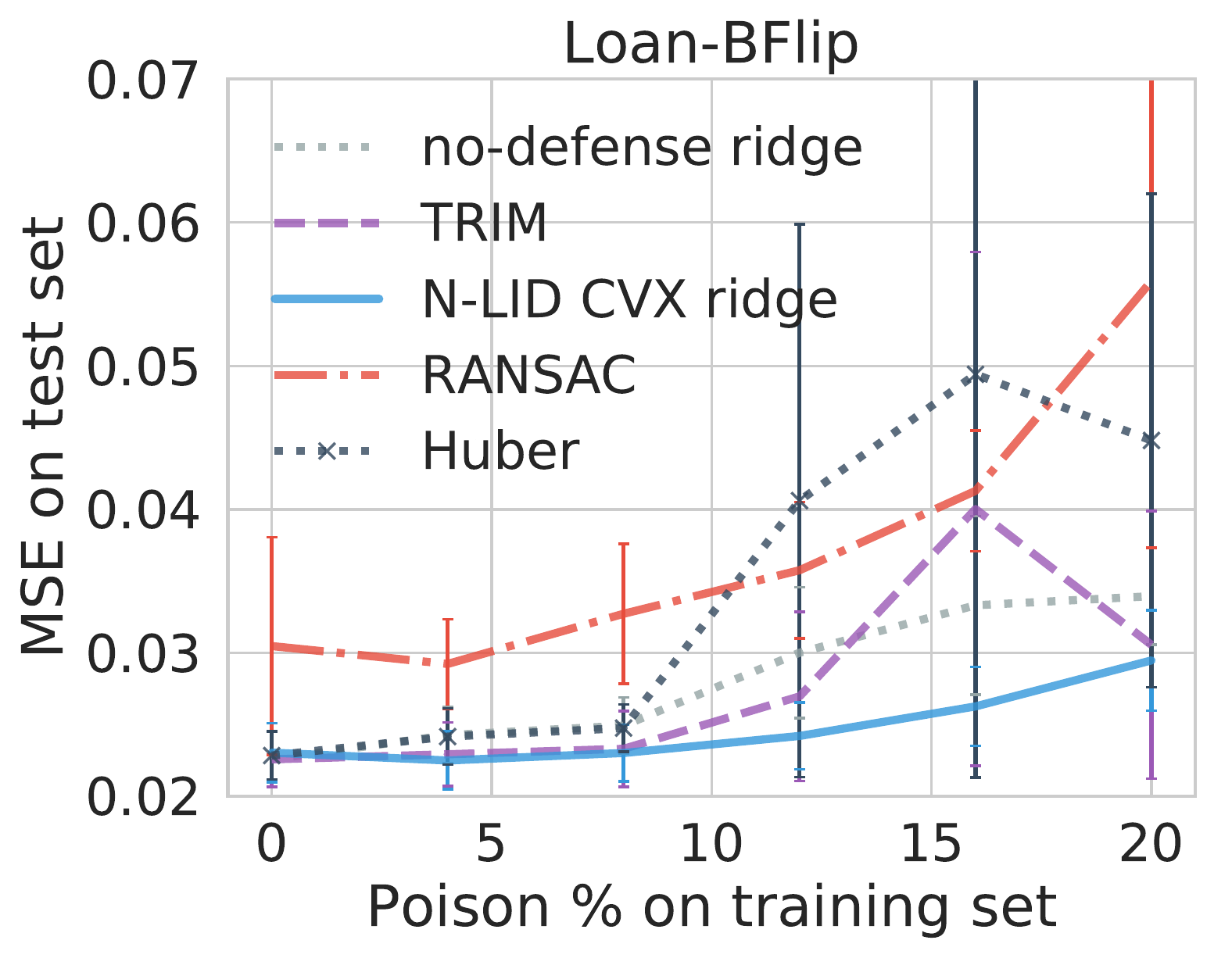}
		\caption{MSE on the Loan dataset.} 
	\end{subfigure}
	\begin{subfigure}[t]{0.4\textwidth} 
		\includegraphics[width=\textwidth]{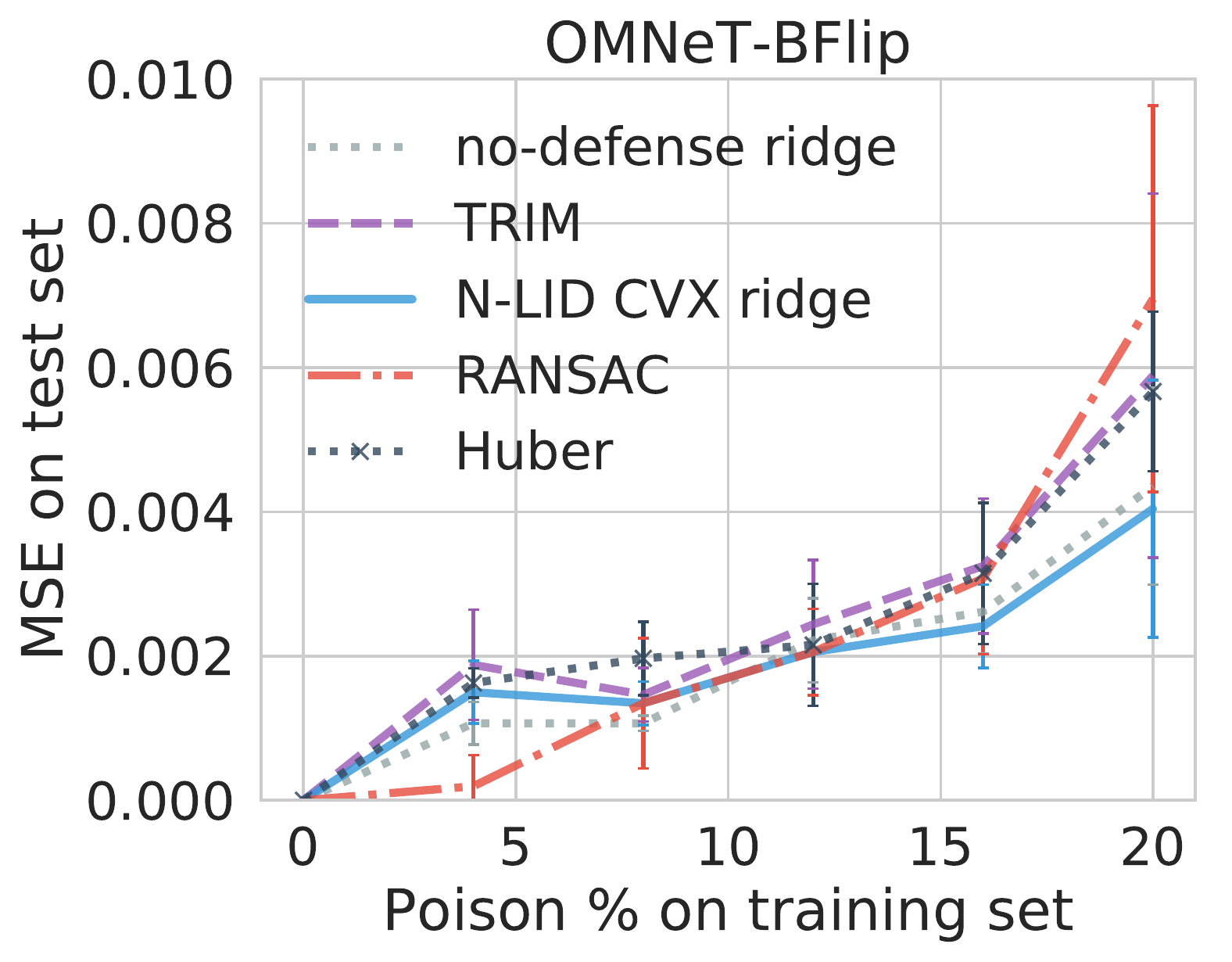}
		\caption{MSE on the OMNeT dataset.} 
	\end{subfigure}
	\begin{subfigure}[t]{0.4\textwidth} 
		\includegraphics[width=\textwidth]{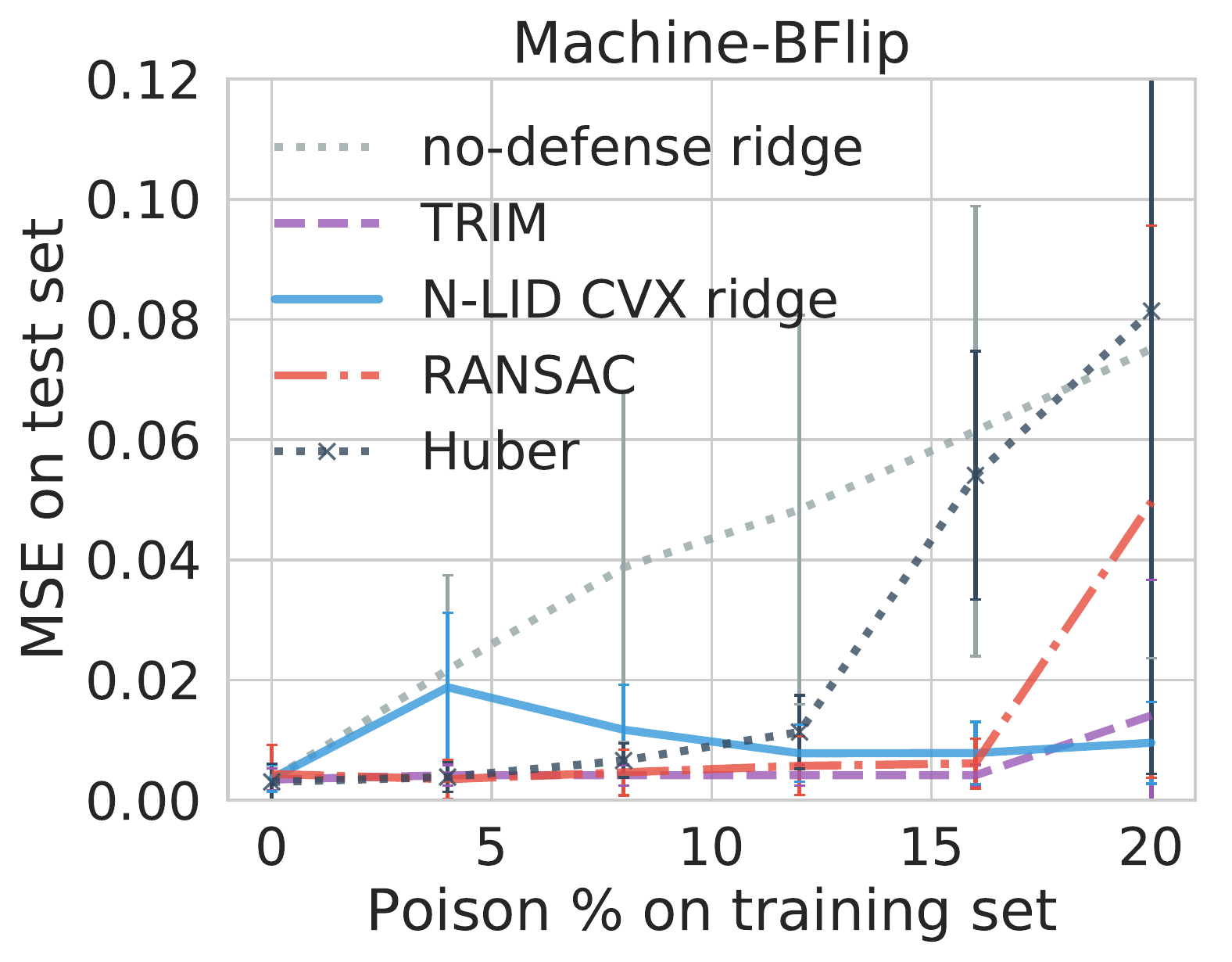}
		\caption{MSE on the Machine dataset.} 
	\end{subfigure}
	\caption{The average MSE on the test set for each defense algorithm under the B-Flip Opt attack. The poisoning rate is increased from $0\%$ to $20\%$.}
	\label{fig:results_3_datasets}
\end{figure*}

\subsubsection{In the absence of an attack}
As stated in Section \ref{sec:reg_attack_model}, we assume that the learner is unaware of the attacker; therefore the training data may or may not be manipulated by the adversary. Hence it is possible to have a scenario where the learner utilizes a defense algorithm when an attack is not present.

It is common for adversarial learning algorithms to sacrifice performance in the absence of an attack to improve their attack resilience in the presence of attacks. Improving attack resilience often results in decreased performance in the absence of an attack and vice versa. However, as Figure \ref{fig:results_3_datasets} shows, all the adversarial/robust learners considered in the experiments perform similarly to a learner with no built-in defense when the poisoning rate is zero. This result may be explained by the fact that there is a high level of redundancy in the linear regression datasets. Therefore even if a learner removes/de-emphasizes a subset of samples during training even when there is no attack, the remaining data samples are sufficient to obtain an accurate estimator. 

\begin{table*}[]
	\centering
	\caption{The percentage increase/decrease of average MSE of each defense algorithm compared to a ridge regression model with no defense. The best results are indicated in \textbf{bold} font. The rows corresponding to the defense algorithms introduced in this work are highlighted in grey.}
	\label{tab:avg_error_rates}
	\begin{tabular}{lrrrrrrrrrr}
		\toprule
		& \multicolumn{2}{c}{House}             & \multicolumn{2}{c}{Loan}             & \multicolumn{2}{c}{OMNeT}            & \multicolumn{2}{c}{Grid}           & \multicolumn{2}{c}{Machine}             \\
		& BFlip             & IFlip             & BFlip             & IFlip            & BFlip            & IFlip            & BFlip             & IFlip             & BFlip             & IFlip             \\\midrule
		\rowcolor{Gray}
		N-LID LR ridge  & \textbf{-87.27\%} & \textbf{-84.24\%} & -7.53\%           & -4.78\%          & \textbf{-20.53\%} & \textbf{-29.36\%} & -4.29\%          & -0.86\%          & -80.64\%          & -71.63\%          \\
		\rowcolor{Gray}
		N-LID CVX ridge & -66.83\%          & -67.74\%          & \textbf{-12.16\%} & \textbf{-8.55\%} & 0.30\%            & -3.52\%           & -2.84\%          & -0.13\%          & -76.12\%          & -76.63\%          \\
		TRIM            & -38.13\%          & -70.43\%          & -1.59\%           & -0.72\%          & 31.76\%           & 6.83\%            & -6.74\%          & -0.97\%          & \textbf{-86.24\%} & -61.96\%          \\
		RANSAC          & 53.26\%           & 61.83\%           & 33.36\%           & 22.63\%          & 20.36\%           & 61.64\%           & -5.92\%          & -1.05\%          & -70.24\%          & \textbf{-83.37\%} \\
		Huber           & 99.81\%           & 46.97\%           & 22.22\%           & 23.93\%          & 28.61\%           & 7.24\%            & \textbf{-7.18\%} & \textbf{-2.14\%} & -35.55\%          & -39.60\%          \\
		\bottomrule 
	\end{tabular}
\end{table*}

\begin{table*}[ht]
	\centering
	\caption{The average training time (s) as a factor of the training time of a ridge regression model with no defense. The best results are indicated in \textbf{bold} font. The rows corresponding to the defense algorithms introduced in this work are highlighted in grey.}
	\label{tab:running_time}
	\begin{tabular}{lrrrrrrrrrr}
		\toprule
		& \multicolumn{2}{c}{House}                             & \multicolumn{2}{c}{Loan}                              & \multicolumn{2}{c}{OMNeT}                              & \multicolumn{2}{c}{Grid}                           & \multicolumn{2}{c}{Machine}                             \\
		& \multicolumn{1}{l}{BFlip} & \multicolumn{1}{l}{IFlip} & \multicolumn{1}{l}{BFlip} & \multicolumn{1}{l}{IFlip} & \multicolumn{1}{l}{BFlip} & \multicolumn{1}{l}{IFlip} & \multicolumn{1}{l}{BFlip} & \multicolumn{1}{l}{IFlip} & \multicolumn{1}{l}{BFlip} & \multicolumn{1}{l}{IFlip} \\\midrule
		\rowcolor{Gray}
		N-LID LR ridge  & 14.85                     & 17.31                     & 47.36                     & 42.21                     & 30.46                     & 28.08                     & 127.13                    & 88.98                     & 199.11                    & 198.27                   \\
		\rowcolor{Gray}
		N-LID CVX ridge & \textbf{11.30}            & \textbf{12.88}            & \textbf{33.83}            & \textbf{30.04}            & \textbf{22.91}            & \textbf{19.52}            & 86.43                     & 60.85                     & \textbf{9.89}             & \textbf{9.78}             \\
		TRIM            & 19.55                     & 23.32                     & 39.74                     & 41.87                     & 33.87                     & 31.47                     & 83.83                     & 90.47                     & 15.49                     & 11.04                     \\
		RANSAC          & 35.65                     & 40.31                     & 74.62                     & 68.32                     & 41.48                     & 39.28                     & 124.35                    & 73.19                     & 128.26                    & 95.18                     \\
		Huber           & 122.18                    & 149.28                    & 182.52                    & 167.81                    & 207.86                    & 169.31       			  & \textbf{23.24}            & \textbf{11.96}            & 19.50                     & 15.08                     \\
		\bottomrule             
	\end{tabular}
\end{table*}

\subsubsection{In the presence of an attack}
First, we consider the performance of the undefended ridge learner, as shown in Figure \ref{fig:results_3_datasets}. We observe that the MSE values increase by $25.6\%$ and $14.56\%$ on the House dataset, $0.50\%$ and $0.27\%$ on the Loan dataset, $0.46\%$ and $0.01\%$ on the Grid dataset, and $20.75\%$ and $20.15\%$ on the Machine dataset under BFlip and IFLip Opt attacks respectively. We observe a similar trend in the OMNeT dataset, as well. These results suggest that (i) undefended regression learners can be severely affected by poisoning attacks, and (ii) among the two attack configurations considered, BFlip is more potent than IFlip as the initial movement of the response variable $y$ is more aggressive compared to IFlip.

We now compare the performance of an N-LID LR weighted ridge model and N-LID CVX weighted ridge model against TRIM \cite{jagielski2018manipulating}, RANSAC \cite{fischler1981random} and Huber \cite{huber1992robust} under poisoning attacks. Table \ref{tab:avg_error_rates} compares their average MSE values with the average MSE of a ridge regression model with no defense (shown as the percentage increase or decrease).

First, we consider the performance of the defenses on the three datasets with a relatively large number of dimensions (i.e., House, Loan, and OMNeT). Considering the experimental evidence on the performance of the robust learners (i.e., Huber and RANSAC), we see that they are not effective at defending against poisoning attacks when the dimensionality of the data is high. In fact, their performance is significantly worse than an undefended ridge regression model. The average MSE of RANSAC is up to $61.83\%$ higher on the House dataset, up to $33.36\%$ higher on the Loan dataset and up to $61.64\%$ higher on the OMNeT dataset when compared with an undefended ridge model. Huber also increases the average MSE by $99.81\%$, $23.93\%$, and $28.61\%$, respectively, for the three datasets. This finding is consistent with that of Jagielski et al. \cite{jagielski2018manipulating} who reported similar performance for robust learners when used with high-dimensional datasets.

This behavior of Huber and RANSAC may be due to high dimensional noise. In high dimensional spaces, poisoned data might not stand out due to the presence of high dimensional noise \cite{xu2012outlier}, thereby impairing their performance. Moreover, it should be noted that these defenses are designed against stochastic noise/outliers, not adversarially poisoned data. In an adversarial setting, a sophisticated adversary may poison the data such that the poisoned samples have a similar distribution to normal data, thereby, reducing the possibility of being detected.

In contrast, TRIM, a defense that is designed against poisoning attacks, has up to $70.43\%$ and $1.59\%$ lower MSE values compared to an undefended ridge model on the House and Loan datasets. Interestingly, on the OMNeT dataset, we observe $31.76\%$ and $6.8\%$ higher MSE values for TRIM. Readers may also notice that the prediction performance of TRIM shown in \cite{jagielski2018manipulating} is different from what we have obtained. Analyzing the provided code revealed that in their experiments the poisoned training dataset is created by appending the poisoned samples $\bar{X}$ to the matrix $X$ containing the pristine samples (resulting in $\tilde{X}\in \mathbb{R}^{(n+p)\times d}$). During the first iteration of TRIM, the first $n$ rows are selected to train the regression model, which happens to be the pristine data in this case. To avoid this issue, in our experiments, we randomly permute the rows of $\tilde{X}$ before training with TRIM.

The N-LID based defense mechanisms consistently outperformed the other defenses against all the attack configurations on the three datasets with a large number of dimensions. On the House dataset, N-LID LR weighted ridge model has the best prediction performance ($87.27\%$ lower MSE) with the N-LID CVX weighted ridge model being closely behind ($67.74\%$ lower MSE). A similar trend is present on the OMNeT dataset, where the N-LID LR weighted ridge model has a $29.36\%$ lower average MSE. On the Loan dataset, we observe that N-LID CVX weighted ridge model has the best performance with a $12.16\%$ lower average MSE. It is interesting to note that the performance of the N-LID CVX weighting mechanism, which makes no assumptions of the attack, is on par with the N-LID LR baseline weighting scheme in the majority of the test cases considered.

Turning now to the experimental evidence on the performance of the defenses on the two datasets with a small number of dimensions (i.e., Grid and Machine), we see that all the defenses considered can outperform the ridge regression learner without a defense. What stands out in Table \ref{tab:avg_error_rates} is that the robust regression learners (Huber and RANSAC) that performed worse than an undefended ridge regression model on the three datasets with a large number of dimensions, have significantly lower MSE values on the two datasets with the smaller number of dimensions. Huber has the best prediction performance on the Grid dataset even exceeding the N-LID based defenses. It has up to $7.18\%$ and $39.60\%$ lower MSE values, respectively, for the two datasets. The average MSE value of RANSAC is $5.92\%$ lower on the Grid dataset while having an $83.37\%$ lower MSE on the Machine dataset. 

\begin{figure}[ht]
	\centering
	\begin{subfigure}[t]{0.4\textwidth} 
		\includegraphics[width=\textwidth]{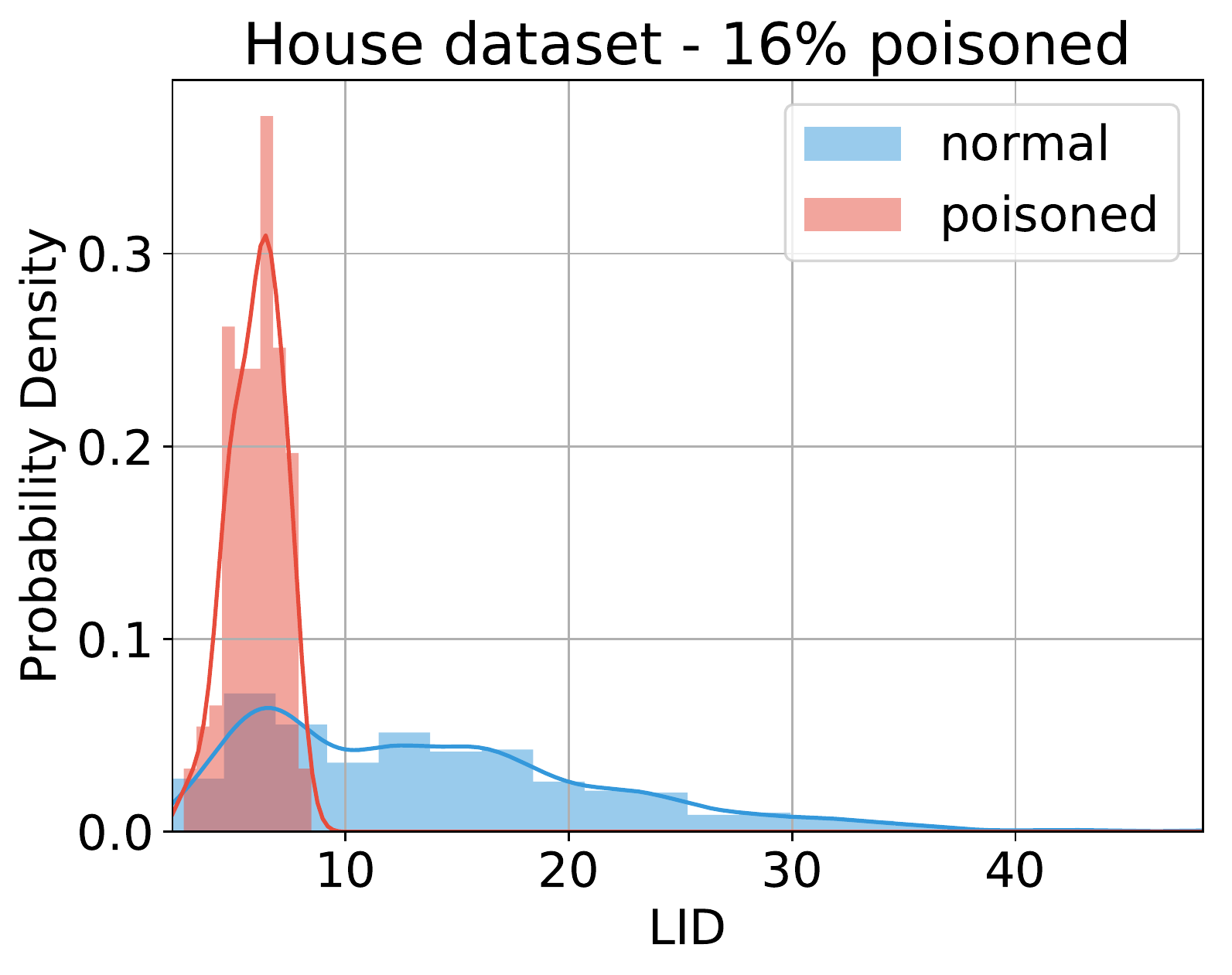}
		\caption{LID distributions in House dataset.} 
		\label{fig:lid_graph_house}
	\end{subfigure}
	\begin{subfigure}[t]{0.4\textwidth} 
		\includegraphics[width=\textwidth]{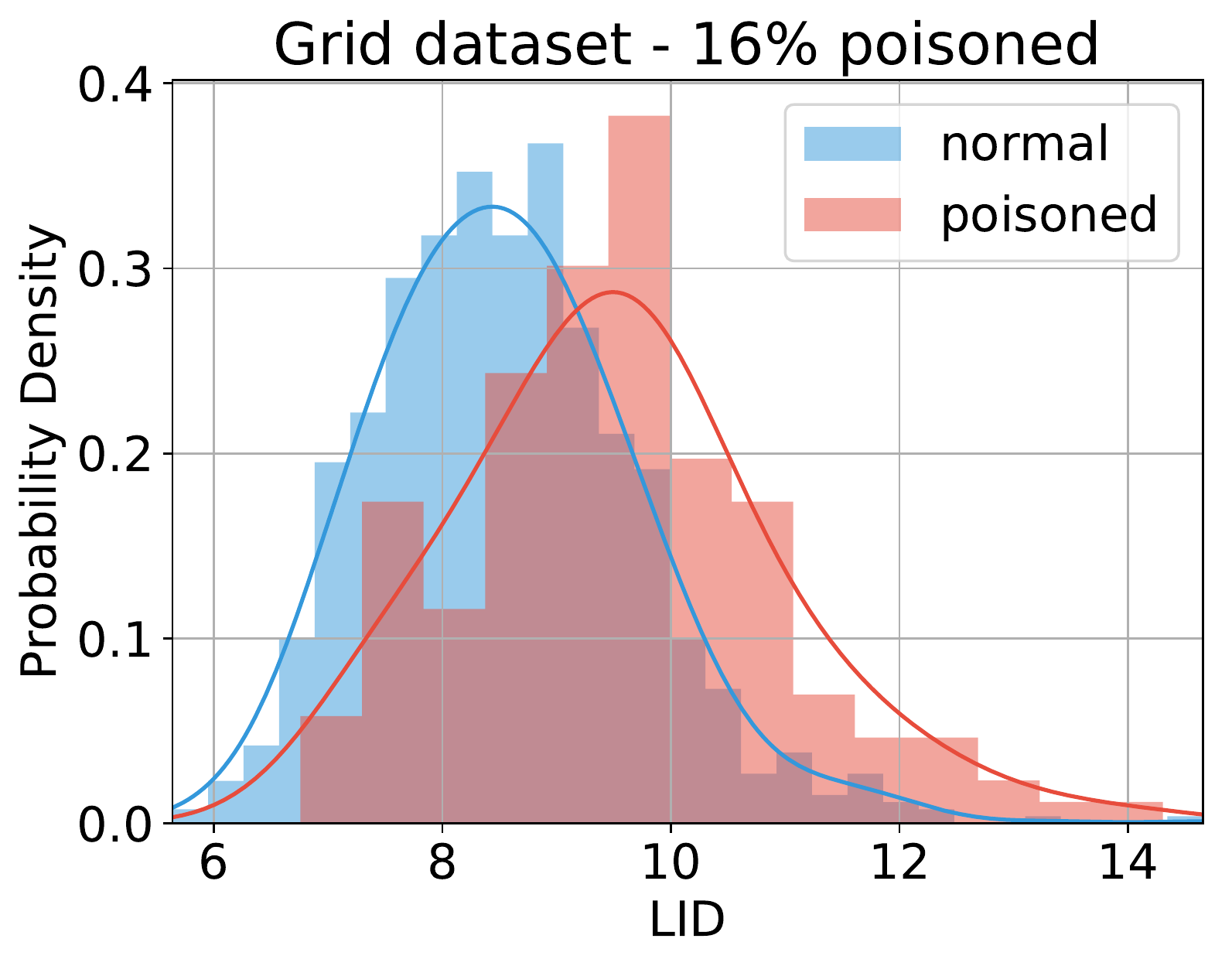}
		\caption{LID distributions in Grid dataset.} 
		\label{fig:lid_graph_grid}
	\end{subfigure}
	\caption{The LID distribution of poisoned and normal samples when 16\% of the training data is poisoned.}
	\label{fig:n_lid_dist}
\end{figure}

N-LID based defenses perform consistently across all the datasets considered when compared against an undefended ridge regression model. However, their performance improvement is relatively smaller on Grid and Machine. The lack of performance on the two datasets with a small number of dimensions may be explained by Figure \ref{fig:n_lid_dist}, which shows the LID (not N-LID) distributions of normal and poisoned samples in two datasets with different dimensions. The LID estimate of a sample is the dimension of the true low-dimensional subspace (at least approximately) in which the particular sample lies (Section \ref{sec:lid_theory}). In other words, it describes the optimal number of features needed to explain the salient features of the dataset. For datasets that have a small number of features, the range of values LID could take is also narrow (as LID is less than the actual number of dimensions). Therefore, if the majority of the features of a particular dataset are salient features, the resulting LID distributions of normal and poisoned samples would have a large overlap (Figure \ref{fig:lid_graph_grid}). Overlapping LID distributions lead to overlapping N-LID distributions, thereby impairing the performance of LID based defenses on such datasets.

In contrast, datasets with a large number of dimensions may allow for a wider range of LID values. In such high-dimensional spaces poisoned and normal samples would exhibit distinguishable LID distributions as shown in Figure \ref{fig:lid_graph_house} (unless a majority of the features are salient features as explained above). Moreover, in such spaces where most data points seem equidistant, the N-LID based defenses have good performance because they characterize poisoned samples from normal samples using the true low-dimensional manifold of the data. However, the issue of obtaining distinguishable LID distributions for poisoned and normal samples for datasets with a high percentage of salient features is an intriguing one that could be usefully explored in further research.


\subsubsection{Practical implications of poisoning attacks} 
We now turn to the practical implications of the poisoning attacks on the case study considered in this paper. In a security application such as the one discussed, having the decision variable change by even a small amount could have dire consequences. We observe that the BFlip attack changes the assigned risks of $10.78\%$ of the transmission sources by $50\%$ when a ridge regression model without a defense is used (at $20\%$ poisoning rate). This means a transmission source that should ideally be assigned a high-risk probability may be assigned a significantly lower risk value or vice versa. Nearly half of all the transmission sources ($45.51\%$) have their assigned risks altered by at least $10\%$. However, the N-LID LR defense model resulted in none of the transmission sources having their assigned risk increased by over $10\%$.

\begin{figure}[ht]
	\centering
	\begin{subfigure}[t]{0.4\textwidth} 
		\includegraphics[width=\textwidth]{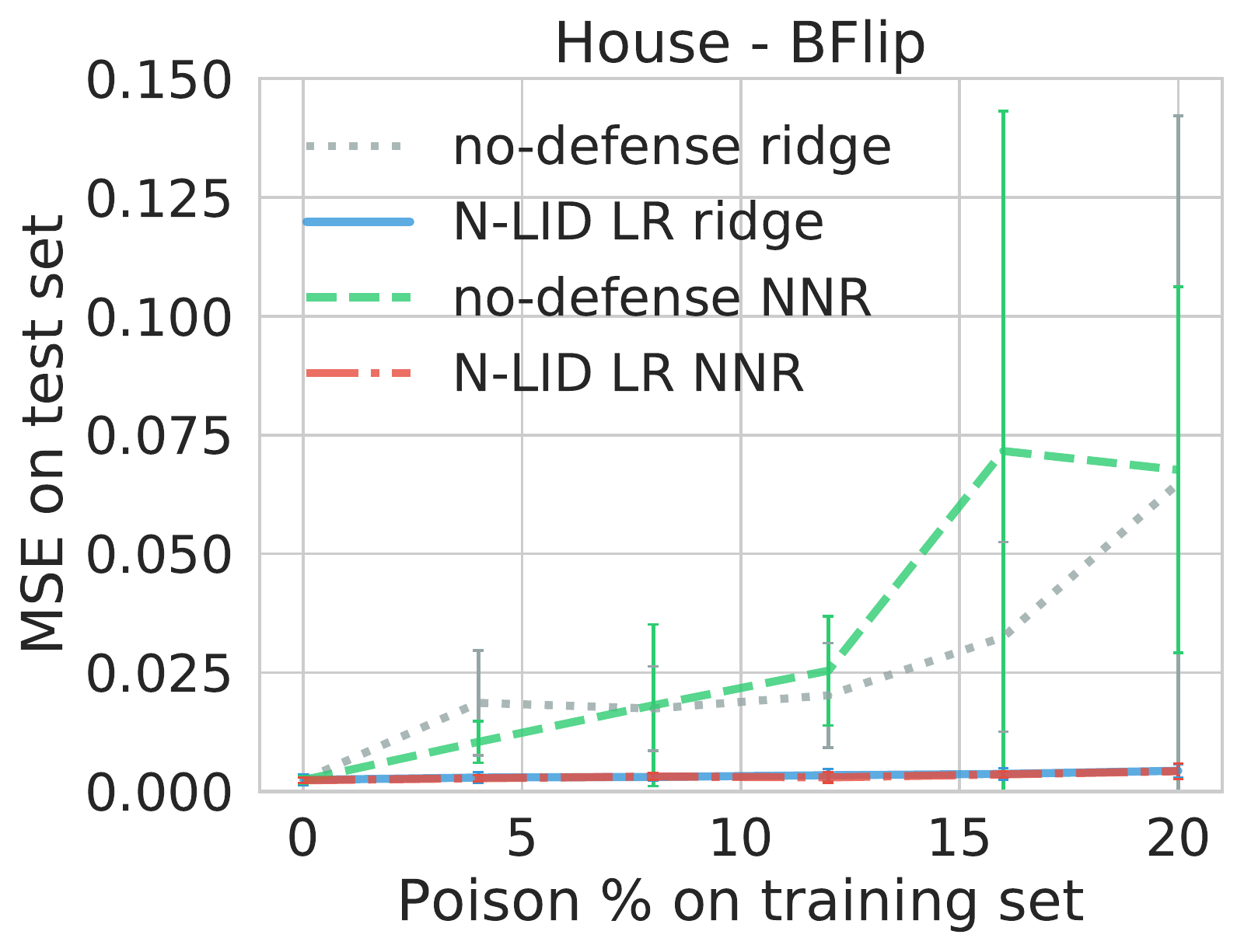}
		\caption{MSE on the House dataset.} 
	\end{subfigure}
	\begin{subfigure}[t]{0.4\textwidth} 
		\includegraphics[width=\textwidth]{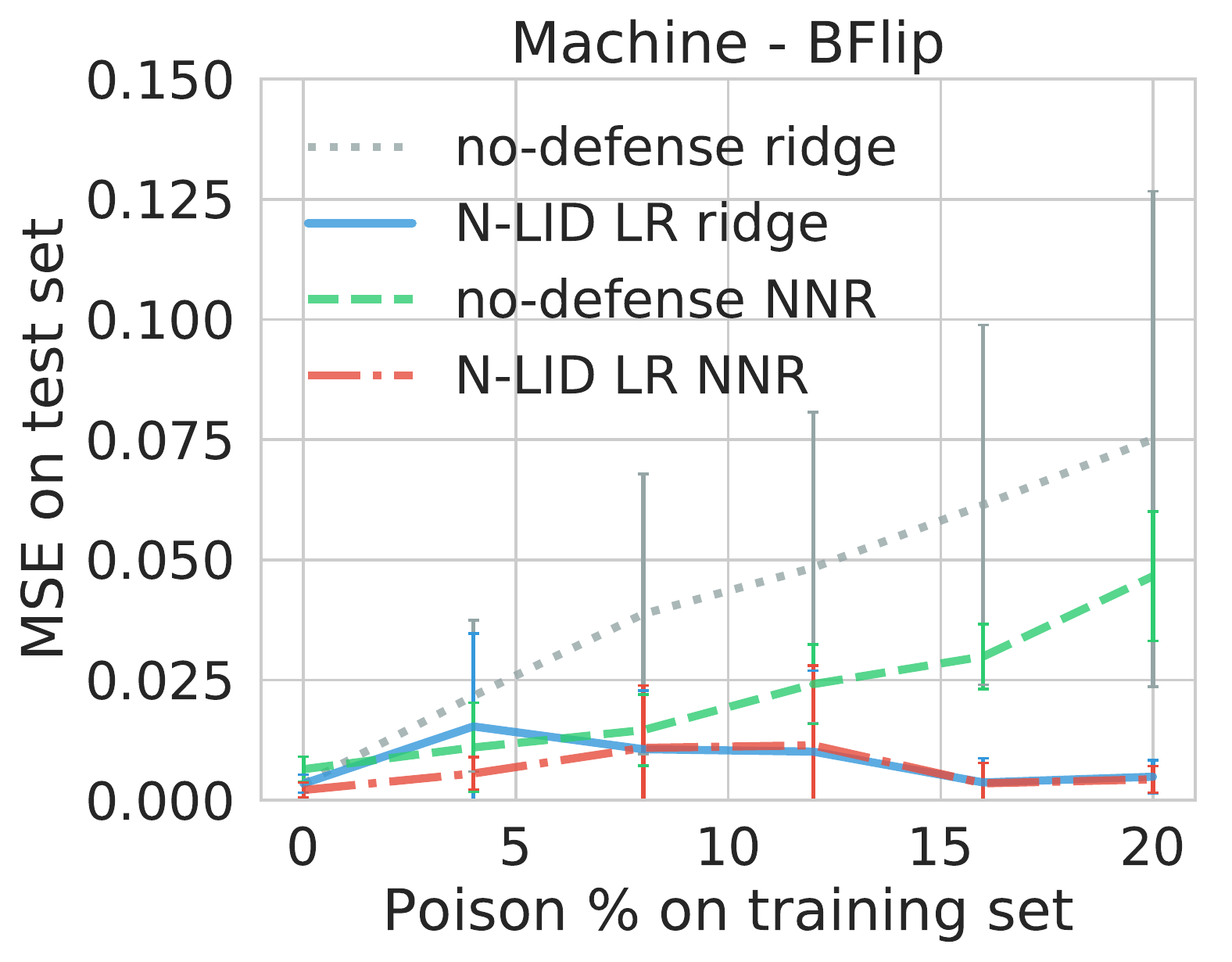}
		\caption{MSE on the Machine dataset.} 
	\end{subfigure}
	\caption{The average MSE of NNR and ridge models when the poisoning rate is increased from $0\%$ to $20\%$.}
	\label{fig:results_transferability}
\end{figure}

\subsubsection{Computational complexity} 
Table \ref{tab:running_time} compares the average training time (in seconds) for the learners considered in the experiments. Because ridge regression with no defense has the lowest training time, we report the training times of the other algorithms as a factor of its training time. It is apparent from this table that N-LID CVX ridge has the lowest training time in eight of the ten experiments considered. Its training times are only $11.30$ times higher on the House dataset, $33.83$ times higher on the Loan dataset, $86.43$ times higher on the Grid dataset, $9.89$ times higher on the Machine dataset, and $22.91$ times higher on the OMNeT dataset. TRIM, being an iterative method, has higher training times compared to N-LID CVX ridge in the majority of the test cases. But its computational complexity is significantly lower compared to RANSAC, which is another iterative method.

Of the learners considered, Huber has the most variation in average training time. We observe that on the three datasets with a large number of dimensions, its training times are up to $207.86$ times higher compared to an undefended ridge model. But on the two datasets with a small number of dimensions, its training times are significantly less (with $11.96$ being the lowest among all the defenses on the Grid dataset).

The main advantage of the N-LID based methods is that they are not iterative; the N-LID calculation, albeit expensive, is only done once. Therefore they exhibit significant savings in computation time while incurring little or no loss in prediction accuracy. Given the results of the datasets with a small number of dimensions (i.e., machine and grid), the N-LID based methods also appear to be invariant to the number of dimensions of the data, unlike the robust regression methods we have considered.

Note that the computational complexity of N-LID would scale poorly to the number of training data samples due to its $k$-nearest neighbor based calculation. However, the N-LID calculation can be made efficient by estimating the LID of a particular data sample within a randomly selected sample of the entire dataset. We refer the readers to the work of Ma et al. \cite{LID_sarah_ICLR} for further information about this process known as minibatch sampling. Furthermore, calculating the LIDs within each minibatch can be parallelized for improved performance.

\subsubsection{Transferability of the defense} 
We now investigate the transferability of the proposed defense by applying it to a fundamentally different learner, neural network regression (NNR). Neural network based regression models follow a gradient-based optimization algorithm similar to Ridge regression that minimizes the MSE between the model prediction and the correct response variable \cite{NNR}. The algorithm uses a linear activation function for the output layer with a single neuron and backpropagation to compute the gradients. The sample weight-based defense can be used with NNR models by multiplying the loss of each sample by its corresponding weight to increase or decrease its importance in the optimization process.

First, we run the Opt attack against ridge regression and obtain a poisoned dataset. Then we train an NNR model on the poisoned data. The average MSE of the NNR model with no defense is $25.34\%$ higher on the House dataset, $12.29\%$ higher on the Loan dataset, $5.61\%$ higher on the OMNeT dataset, $19.11\%$ higher on the Grid dataset, and $46.67\%$ lower on the Machine dataset compared to a ridge regression model with no defense. This result may be explained by the fact that an NNR model (which is a nonlinear estimator) is too complex for a linear regression problem; therefore, it overfits the perturbed training data, resulting in poor prediction performance on an unperturbed test set.

However, our aim is to demonstrate that our proposed defense can be applied to different learning algorithms, not to compare the linear ridge regression model with a NNR model. To that end, we observe that if the loss function of the NNR is weighted using the N-LID LR weighting scheme we introduce, it can maintain a near $0\%$ increase in MSE when the poisoning rate is increased to $20\%$. These findings demonstrate the flexibility of our defense mechanism.

In summary, these results show that (i) robust learners are unable to withstand carefully crafted adversarial attacks on datasets where the number of dimensions is large, whereas adversarial learners can, (ii) N-LID based defenses that de-emphasize the effects of suspected samples consistently perform better in terms of lower MSE values and running times compared to other defenses, (iii) although N-LID CVX makes no assumptions of the attack, its performance is comparable to the baseline N-LID LR defense, and (iv) the N-LID based defenses can successfully be used in conjunction with different algorithms such as NNR models.

\section{Conclusions}\label{sec:regression_conclusions}
This paper addresses the problem of increasing the attack resistance of linear regression models against data poisoning attacks. We observed that carefully crafted attacks could significantly degrade the prediction performance of regression models, and robust regression models are unable to withstand targeted adversarial attacks on datasets with a large number of dimensions. We introduced a novel LID measure that took into account the LID values of a data sample's neighbors and introduced several weighting schemes that alter each sample's influence on the learned model. Experimental results suggest that the proposed defense mechanisms are quite effective and significantly outperform prior art in terms of accuracy and computational costs. Further research should be undertaken to investigate the effects of poisoning attacks on non-linear regression models and how N-LID would perform in such scenarios.

%
\bibliographystyle{IEEEtran}
\bibliography{references}
\newpage

\end{document}